\documentclass[11pt]{article}

\usepackage[preprint]{acl}

\usepackage{times}
\usepackage{latexsym}

\usepackage[T1]{fontenc}

\usepackage[utf8]{inputenc}

\usepackage{microtype}

\usepackage{inconsolata}

\usepackage{graphicx}

\usepackage{graphicx} %
\usepackage{natbib}  %
\usepackage{caption} %
\usepackage{algorithm}
\usepackage{listings}
\usepackage{algorithmic}
\usepackage{amsmath}
\usepackage{booktabs}
\usepackage{multirow}
\usepackage[outdir=./]{epstopdf}
\usepackage{enumitem}
\usepackage{caption}
\usepackage{subcaption}
\usepackage{subfloat}
\usepackage{newfloat}
\usepackage{graphicx}
\usepackage{svg} 
\usepackage[normalem]{ulem}
\usepackage{framed}
\usepackage{mdframed}
\usepackage{xcolor}
\usepackage{lipsum}
\usepackage{float}
\usepackage{hyperref}
\usepackage{tabularx}
\usepackage{multirow}

\usepackage[most]{tcolorbox}
\tcbuselibrary{breakable}

\definecolor{shadecolor}{gray}{0.9}

\newlist{todolist}{itemize}{2}
\setlist[todolist]{label=$\square$}
\usepackage{pifont}

\usepackage{array}
\newcolumntype{L}[1]{>{\raggedright\let\newline\\\arraybackslash\hspace{0pt}}m{#1}}
\newcolumntype{C}[1]{>{\centering\let\newline  \\\arraybackslash\hspace{0pt}}m{#1}}
\newcolumntype{R}[1]{>{\raggedleft\let\newline \\\arraybackslash\hspace{0pt}}m{#1}}

\AtBeginDocument{%
  \providecommand\BibTeX{{%
    \normalfont B\kern-0.5em{\scshape i\kern-0.25em b}\kern-0.8em\TeX}}
}
\begin{document}
\title{Rationality Check! Benchmarking  the Rationality of Large Language Models}

\author{\\
\textbf{Zhilun Zhou}\textsuperscript{1\thanks{These two authors contributed equally.}} \quad
\textbf{Jing Yi Wang}\textsuperscript{1\footnotemark[1]}\quad
\textbf{Nicholas Sukiennik}\textsuperscript{1} \\
\textbf{Chen Gao}\textsuperscript{1\thanks{Corresponding authors.} } \quad
\textbf{Fengli Xu}\textsuperscript{1} \quad
\textbf{Yong Li\textsuperscript{1\footnotemark[2]}}\quad
\textbf{James Evans}\textsuperscript{2,3} \\
\textsuperscript{1}Department of Electronic Engineering, BNRist, Tsinghua University \\
\textsuperscript{2}Department of
Sociology, University of Chicago, Chicago, IL, USA\\
\textsuperscript{3}Santa Fe Institute, Santa Fe, NM, USA\\
\{zzl22, jy-w22\}@mails.tsinghua.edu.cn
}



\maketitle
\begin{abstract}
Large language models (LLMs), a recent advance in deep learning and machine intelligence, have manifested astonishing capacities, now considered among the most promising for artificial general intelligence. With human-like capabilities, LLMs have been used to simulate humans and serve as AI assistants across many applications. As a result, great concern has arisen about whether and under what circumstances LLMs think and behave like real human agents. Rationality is among the most important concepts in assessing human behavior, both in thinking (\textit{i.e.}, theoretical rationality) and in taking action (\textit{i.e.}, practical rationality). In this work, we propose the first benchmark for evaluating the omnibus rationality of LLMs, covering a wide range of domains and LLMs. The benchmark includes an easy-to-use toolkit, extensive experimental results, and analysis that illuminates where LLMs converge and diverge from idealized human rationality. We believe the benchmark can serve as a foundational tool for both developers and users of LLMs.
Our code and dataset can be found at \url{https://github.com/tsinghua-fib-lab/LLM-Rationality-Benchmark}
\end{abstract}
\maketitle

\section{Introduction}\label{sec::intro}




Large Language Models (LLMs)~\cite{zhao2023survey} represent a significant advancement in the field of artificial intelligence, manifesting human-like capability in natural language processing, understanding, and generation.
As emphasized in developmental psychology, language is a primary means through which children form mental models of the world~\cite{mccall1977challenges}, constituting one of humanity's fundamental abilities and a primary tool for interaction with the world.
Therefore, besides linguistic abilities, LLMs also exhibit strong performance on complex reasoning and planning tasks, such as solving mathematical problems~\cite{romera2023mathematical}, playing video games~\cite{zhu2023ghost}, role-playing~\cite{shanahan2023role}, etc.
This showcases LLMs' promising potential in many fields as an \textbf{agent} rather than a language model alone~\cite{xi2023rise,wang2023survey}.
As a result, LLM agents have been employed as emulators in many fields to simulate human activities ~\cite{gao2023large} and serve as AI assistants~\cite{boiko2023emergent,huang2023benchmarking}. 
In these contexts, LLMs are utilized in problem-solving and decision-making tasks in place of humans. 
Nevertheless, the decisions made by LLMs bear immense significance for the stability of human society, particularly in critical domains dealing with personal privacy or security.
Insofar as we engaged LLM agents for these tasks, it is critical to ensure they can demonstrate reasonable and responsible beliefs and actions, which motivates our assessment their \textbf{rationality} in this paper.


Rationality is an anchoring concept underlying human decision-making and belief, which can be understood as the quality of being guided by reasons or being reasonable~\cite{knauff2021handbook}. It typically implies an agent has reflected on the probable consequences of their actions and the goals they were designed to realize. In this way, rationality is a complex concept that involves reasoning, values, and even emotions. Any particular domain's evaluations or definitions will only cover specific aspects.
Nevertheless, it is widely acknowledged that rationality can be divided into theoretical and practical aspects~\cite{wahlstrom1999discussion}, with the former focusing on whether the internal belief is reasonable, and the latter emphasizing the capacity to make appropriate decisions in real-world contexts.
These two folds of rationality are closely related and are both quite essential in assessing the final actions. 
For LLMs, an extensive evaluation of twofold rationality involves determining whether they have human-like rationality levels in handling real-world decisions or task execution, thereby establishing a solid foundation for safer and more responsible AI applications.

Despite its importance, extensively benchmarking the rationality of LLMs remains unexplored. In this work, we take the pioneering step to establish a systematic benchmark for measuring the rationality of various LLMs, covering different domains' definitions and assessments of rationality.
Specifically, we organized the framework into six distinct domains: psychology, cognitive science, decision-making, economics, game theory, and collective rationality. 
Leveraging widely acknowledged rationality questionnaires from various domains, we conduct comprehensive assessments of the rationality of various LLMs, which encompass both open-source LLMs and commercial API-based models, aiming to extensively cover commonly used models.
While it might not encompass all individual rationality measures, our benchmark covers the main subclasses of rationality; thus, for an uncovered measure, we believe it easy to find a suitable substitute or reference results in our benchmark.

Overall, our benchmark serves as a fundamental evaluation tool for both developers and users of LLMs. For developers, the benchmark can serve as a reference for optimizing and training models, indicating the position of a model's rationality and identifying potential areas for further enhancement. For users, our benchmark offers a standardized evaluative system, facilitating assessments of rationality levels within specific usage scenarios. This helps mitigate risks associated with applications utilizing LLMs as crucial AI assistance tools in decision-making.

Under our framework for rationality assessment, our experiments have the following findings.
First, LLMs demonstrate an overall level of high rationality, particularly evident in larger-scale models that exhibit greater prowess. 
Second, LLMs show very high collective rationality, showcasing a greater inclination toward cooperation and a more homogeneous AI society compared to human society. 
Third, LLMs showcased higher levels of rationality in non-abstract scenarios, indicating their capability for understanding and making decisions in real-world scenarios. 
Furthermore, the results of our benchmark have revealed a multitude of insightful observations across different domains, including exhibiting both similar and widely different responses to humans.

Our contribution can be summarized as follows.
\begin{itemize}[leftmargin=*]
    \item We propose a comprehensive taxonomy of rationality evaluation based on extensive literature review, which organizes rationality into individual, interpersonal, and societal levels, covering six domains of rationality research. This taxonomy provides a structured foundation for evaluating rationality from both theoretical and practical perspectives.
    \item Based on this taxonomy, we construct the first benchmark for measuring LLMs' rationality, including a wide range of questionnaires, tests, and games.
    \item We conduct a detailed assessment of a bundle of renowned LLMs and rationality scales, with extensive analyses of the results, including comparisons with human rationality, inter-domain analysis, theoretical-practical analysis, the impact of training methodologies and model parameters on rationality, and the relationship between individual and collective rationality. We believe that these conclusions hold significant implications for both developers and users of LLMs. 
    \footnote{Our employed measurements and code can be found at \url{https://github.com/tsinghua-fib-lab/LLM-Rationality-Benchmark}
    }
\end{itemize}

\section{Related Work}\label{sec::related}
\subsection{Evaluation of LLMs}

The most evaluated aspect for LLMs is natural language processing, including natural language understanding~\cite{liang2022holistic,bang2023multitask,pena2023leveraging, yang2023large,lee2023can} and natural language generation~\cite{chang2023survey}.
Additionally, LLM evaluations have expanded to simulations and decision-making contexts, where they act as autonomous agents mimicking human behavior~\cite{gao2023s,argyle2023out,li2023large}.
While these evaluations provide significant insights into LLM capabilities, the evaluation of rationality, a core component of human cognition, remains underdeveloped.
Although some recent studies have started to explored the rationality in LLMs like GPT-4 \cite{macmillan2024ir}, these efforts are limited to test from a single domain of study, e.g., cognitive psychology \cite{macmillan2024ir} or economics \cite{guo2024economics}, providing insufficient data for a comprehensive rationality benchmark. In contrast, our work proposes a more rigorous framework for rationality, systematically assessing LLMs across diverse domains to provide deeper insights into their decision-making processes.

\subsection{Rationality and its evaluation} 

Rationality, defined as being guided by reason, governs both belief formation and decision-making~\cite{knauff2021handbook}. It is often divided into theoretical rationality, which focuses on logical reasoning and belief coherence, and practical rationality, which involves decision-making and mitigating cognitive biases~\cite{wahlstrom1999discussion}. Moreover, rationality is widely studied across disciplines. In psychology, it is examined through constructs such as self-reflection and emotional regulation, assessing an individual’s ability to reflect on thoughts and manage behavior~\cite{shulman1984psychology}. Cognitive science investigates how biases and heuristics affect decision-making~\cite{mccready2014aristotle}. In decision theory, rationality is evaluated based on decision-making styles and biases in uncertain environments~\cite{sleboda2017measurements}. In economics, bounded rationality acknowledges the limitations of human decision-making due to cognitive constraints~\cite{chapman2023econographics}, with constructs like overconfidence and risk attitudes playing a key role~\cite{kahneman2003maps,stango2019we}. Game theory extends rationality to strategic interactions, assessing whether agents can achieve optimal outcomes~\cite{edelman2007internet,graham2017corporate}. Social sciences consider collective rationality, examining group decision-making processes and the “wisdom of the crowd”~\cite{aher2023using,hendrycks2020measuring,hendrycks2021measuring}.

Rationality is commonly assessed through questionnaires, using self-assessments or scenario-based tasks~\cite{zabojnik2004model}. Since human rationality is often evaluated in this way, LLMs can also be assessed by having them respond as participants. Previous studies have applied this approach to evaluate cognitive skills, logical reasoning, and decision-making in LLMs~\cite{binz2023using,betz2020critical,chen2023emergence}. While some research has explored specific aspects of rationality in LLMs~\cite{binz2023using,gao2023s}, no comprehensive benchmark has yet been established. 

\section{Benchmark Construction}

\begin{table*}[h]
\centering
\renewcommand{\arraystretch}{1.13}
\caption{Rationality Measurement and Evaluation}
\vspace{-5px}
\label{rationality-benchmark}
\resizebox{\textwidth}{!}{
\begin{tabular}{|p{2.3cm}|p{3.3cm}|p{3.5cm}|p{4.1cm}|p{7.5cm}|p{5cm}|}
\hline
\textbf{Level} & \textbf{Domain} & \textbf{Rationality Type} & \textbf{Aspect} & \textbf{Assessment Method} & \textbf{Evaluation Metric} \\
\hline

\multirow{9}{*}{\textbf{Individual}}  
& \multirow{3}{*}{Psychology}  
& Theoretical & Self-reflection & Self-Reflection Insight Scale (SRIS)~\cite{grant2002self} & Total score \\ 
\cline{3-6}
& & Practical & Emotion regulation & Emotion Regulation Questionnaire (ERQ)~\cite{gross2003individual} & Average score \\ 
\cline{3-6}
& & Practical & Intrinsic motivation & Need for Cognition (NFC) scale~\cite{cacioppo1984efficient} & Average score \\ 
\cline{2-6}

& \multirow{5}{*}{Cognitive Science}  
& \multirow{2}{*}{Theoretical} & Dual-process reasoning & Rationality-Experiential Inventory (REI)~\cite{pacini1999relation} & Total score \\ 
\cline{5-6}
& & &  & Cognitive Reflection Test (CRT)~\cite{toplak2014assessing} & Accuracy \\ 
\cline{3-6}

& & Theoretical & Logical reasoning & Inductive reasoning~\cite{ekstrom1976kit}, Deductive reasoning~\cite{liu2023logiqa}, Causal reasoning~\cite{waldmann2005seeing,binz2023using} & Accuracy \\ 
\cline{3-6}

& & Theoretical & Contextual reasoning & Defeasible reasoning~\cite{ford2000strategies}, Scientific reasoning~\cite{drummond2017development}, Deontic reasoning~\cite{ragni2017wason} & Accuracy \\ 
\cline{3-6}

& & Practical & Thinking dispositions & Critical Thinking Disposition Scale (CTDS)~\cite{sosu2013development}, Actively Open-Minded Thinking (AOT) Scale~\cite{campitelli2014does} & Average score \\ 
\cline{2-6}

& \multirow{2}{*}{Decision-making}  
& Practical & Decision styles & Decision Style Scale (DSS)~\cite{scott1995decision} & Scores for each decision style\\ 
\cline{3-6}
& & Practical & Decision biases & Availability heuristic~\cite{lichtenstein1978judged}, Base-rate neglect~\cite{erceg2022normative}, Confirmation bias~\cite{rieger2022survey}, Framing effect~\cite{bruine2007individual}, Conjunction fallacy~\cite{burgoyne2023understanding} & Accuracy \\ 
\cline{2-6}

& \multirow{6}{*}{Economics}  
 & Theoretical & Overconfidence & Overconfidence test~\cite{michailova2010development} & Deviation between confidence and performance \\ 
\cline{3-6}
& & Practical & Risk attitude & Domain-Specific Risk Attitude Scale (DOSPERT)~\cite{blais2006domain} & Total risk-taking and risk perception score \\ 
\cline{3-6}
& & Practical & Risk propensity & Risk Propensity Scale~\cite{bachmann2010risk} & Average score \\ 
\cline{3-6}
& & Practical & Loss aversion & Loss Aversion Task~\cite{stango2019we} & Average score \\ 
\cline{3-6}
& & Practical & Time preference & Temporal Discounting Task~\cite{toplak2014assessing,frederick2005cognitive} & Total\\ 
\cline{3-6}
& & Practical & Economic biases & Endowment effect~\cite{kahneman1990experimental}, Sunk cost fallacy~\cite{bruine2007individual}, Gambler’s fallacy~\cite{leonard2016relationship}, Mental accounting~\cite{rieger2022survey}, Regression to the mean~\cite{toplak2014assessing} & Accuracy \\ 
\hline

\multirow{6}{*}{\textbf{Interpersonal}}  
& \multirow{6}{*}{Game Theory}  
& Practical & Auction strategy & Second-price auction~\cite{kahneman2003maps} & Deviation of bid from true value \\ 
\cline{3-6}
& & Practical & Iterative reasoning & Beauty contest~\cite{arthur1991designing} & Closer choice to Nash equilibrium \\ 
\cline{3-6}
& & Practical & Strategic cooperation & One-shot prisoner’s dilemma~\cite{hendrycks2020measuring} & Defection rate \\ 
\cline{3-6}
& & Practical & Repeated cooperation & Finitely repeated prisoner’s dilemma~\cite{hendrycks2020measuring} & Average defection rate \\ 
\cline{3-6}
& & Practical & Public goods provision & One-shot public goods game~\cite{hendrycks2020measuring} & Percentage of private tokens retained \\ 
\cline{3-6}
& & Practical & Long-term public goods provision & Finitely repeated public goods game~\cite{hendrycks2020measuring} & Average percentage of private tokens retained \\ 
\hline

\multirow{2}{*}{\textbf{Societal}}  
& \multirow{2}{*}{Collective Rationality}  
& Practical & Cooperation and coordination & Infinitely repeated prisoner’s dilemma, Battle of the sexes, Minimum effort, Stag hunt~\cite{aher2023using} & Efficiency score \\ 
\cline{3-6}
& & Practical & Collective decision-making & Wisdom of crowds~\cite{aher2023using,zhang2024exploring,hendrycks2020measuring} & Group decision accuracy \\ 
\hline

\end{tabular}
}
\vspace{-10px}
\end{table*}

Rationality is inherently a multidisciplinary concept, and few existing studies offer a comprehensive summary of all its forms. Building on an extensive review of research across various fields~\cite{mele2004oxford,knauff2021handbook}, our benchmark evaluates rationality at three levels: individual, interpersonal, and societal. Each level includes multiple domains and a range of aspects used to assess rationality, as summarized in Table~\ref{rationality-benchmark}.
Notably, the aspects within these domains are categorized into theoretical and practical rationality based on their intrinsic nature. Theoretical rationality focuses on the reasonableness of internal beliefs, while practical rationality emphasizes decision-making in real-world contexts.
In this section, we briefly introduce the benchmark construction process for each domain. More details and examples of questions and prompts are presented in Appendix~\ref{app:measurement_evaluation}. 

\subsection{Psychology}
This domain evaluates rationality from a psychological perspective, covering three key aspects: self-reflection, emotion regulation, and intrinsic motivation. We assess these aspects using well-established psychological questionnaires developed in prior human studies, including the Self-Reflection and Insight Scale (SRIS)~\cite{grant2002self}, the Emotion Regulation Questionnaire (ERQ)~\cite{gross2003individual}, and the Need for Cognition (NFC) scale~\cite{cacioppo1984efficient}. These questionnaires are designed to capture both theoretical rationality (e.g., awareness and clarity of thought) and practical rationality (e.g., emotional control and motivation for thinking). We prompt LLMs to respond as participants to these questionnaires and compute normalized rationality scores based on their responses.

\subsection{Cognitive Science}
This domain assesses rationality by examining LLMs’ cognitive processing, reasoning abilities, thinking dispositions, and susceptibility to biases. Specifically, we evaluate theoretical rationality through dual process theory~\cite{evans2003two,stanovich2000individual}, logical and context-based reasoning (including inductive, deductive, causal, defeasible, scientific, and deontic reasoning)\cite{goswami2010inductive,binz2023using,ragni2017wason}, and thinking dispositions such as critical and open-minded thinking\cite{sosu2013development,campitelli2014does}. Practical rationality is evaluated by measuring a range of cognitive biases closely tied to irrational judgment~\cite{kahneman2003maps,berthet2023heuristics}. All assessments are based on established cognitive science questionnaires or reasoning tasks adapted from prior research. LLMs are prompted to complete these tasks as participants, and their responses are scored and normalized to yield rationality scores ranging from 0 (least rational) to 1 (most rational).

\subsection{Decision-making}
This domain assesses rationality through both decision-making styles and susceptibility to cognitive biases. We adopt the framework from Scott and Bruce~\cite{scott1995decision} to evaluate five decision-making styles, with an emphasis on the rational style as indicative of higher rationality. In addition, we examine LLMs' performance across a set of well-established decision-making heuristics and biases, including availability heuristic, base-rate neglect, confirmation bias, conjunction fallacy, framing effect, and others~\cite{berthet2023heuristics, ceschi2019dimensions}. These tasks are adapted from prior psychological and behavioral research~\cite{erceg2022normative, burgoyne2023understanding, toplak2014assessing, west2003probability, bruine2007individual}, and a lower susceptibility to biases is interpreted as reflecting greater rationality.

\subsection{Economics}
In this domain, rationality is framed around utility maximization~\cite{kahneman2003maps} and is evaluated from both theoretical and practical dimensions. We emphasize economic-specific aspects, such as overconfidence, risk and time preferences, and common economic biases. For theoretical rationality, we assess overconfidence by comparing LLMs’ self-rated confidence with their actual performance~\cite{michailova2010development}. For practical rationality, we examine risk preference (via risk attitude, risk propensity~\cite{blais2006domain,bachmann2010risk}, and loss aversion~\cite{stango2019we}), time preference (using temporal discounting tasks~\cite{frederick2005cognitive,toplak2014assessing}), and economic biases, including endowment effect~\cite{franciosi1996experimental}, gambler’s fallacy~\cite{leonard2016relationship}, mental accounting~\cite{rieger2022survey}, regression to the mean~\cite{toplak2014assessing}, and sunk cost fallacy~\cite{bruine2007individual}. All tasks and questionnaires are adapted from established economic psychology literature, and rationality scores are determined based on lower susceptibility to bias and greater alignment with normative economic reasoning.

\subsection{Game Theory} 
This domain assesses rationality at the interpersonal level, where agents must make strategic decisions while anticipating the actions of others. We adopt several classic economic games with well-defined Nash or subgame perfect Nash equilibria (SPNE), including the second-price auction, beauty contest, one-shot and finitely repeated prisoner’s dilemma, and one-shot and finitely repeated public goods games. These games are widely used in behavioral economics to evaluate interactive decision-making. 
In our benchmark, we construct several agents with the same LLM and let them play the games with each other. Rationality is measured by how closely their behaviors align with the equilibrium strategies. 

\subsection{Collective Rationality}
At the societal level, this domain assesses rationality through group behaviors such as cooperation and coordination and the wisdom of crowds. For cooperation and coordination, we evaluate whether LLM agents can work together to achieve efficient outcomes in classic game-theoretic settings, including the infinitely repeated prisoner’s dilemma, battle of the sexes, minimum effort, and stag hunt. Rationality is quantified using an efficiency score~\cite{bednar2012behavioral}, which compares actual agent payoffs to the possible maximum and minimum. For the wisdom of crowds, we investigate whether groups of LLMs can collectively produce more accurate decisions than individuals. Following prior work~\cite{aher2023using, zhang2024exploring}, we assess group performance through answer aggregation and multi-agent debate on general knowledge, multiple-choice~\cite{hendrycks2020measuring}, and math questions~\cite{hendrycks2021measuring}. Higher accuracy in these tasks indicates stronger collective rationality.

\section{Experiment Setup}

\subsection{Toolkit}\label{sec::toolkit} 

We developed a toolkit to test models on the same prompts, querying multiple models (both API and local) simultaneously and reporting performance across different rationality domains. The toolkit manages API and local model requests, retrieves responses asynchronously, and processes results based on the benchmarking scheme for rationality assessment.
Specifically, the toolkit queries commercial API-based models and open-source models using a unified interface for both input and output, enabling seamless comparisons between models of different types.
Then, the toolkit processes model responses according to our rationality benchmarking framework, assessing various domains such as psychology, cognitive science, decision-making, and game theory, and compares results across models in each rationality domain.
Finally, the toolkit generates heatmaps visualizing the performance of each model across all rationality types and compares LLM performance to human benchmarks, providing a comprehensive view of rationality levels among the evaluated models.





\subsection{Models} 
We evaluate a mix of commercial models (e.g.,
GPT-4o, DeepSeek) and open-source models (e.g.,
Llama2, Vicuna) in our benchmark. Commercial
models, accessible via API, typically perform better due to extensive resources, while open-source
models offer versatility and modification opportunities. A complete list of models, including sizes,
is provided in Appendix~\ref{app:models}.


\section{Results}\label{sec::exp}  

\begin{figure}[t!]
    \centering
    \subfloat{\includegraphics[width=0.96\linewidth]{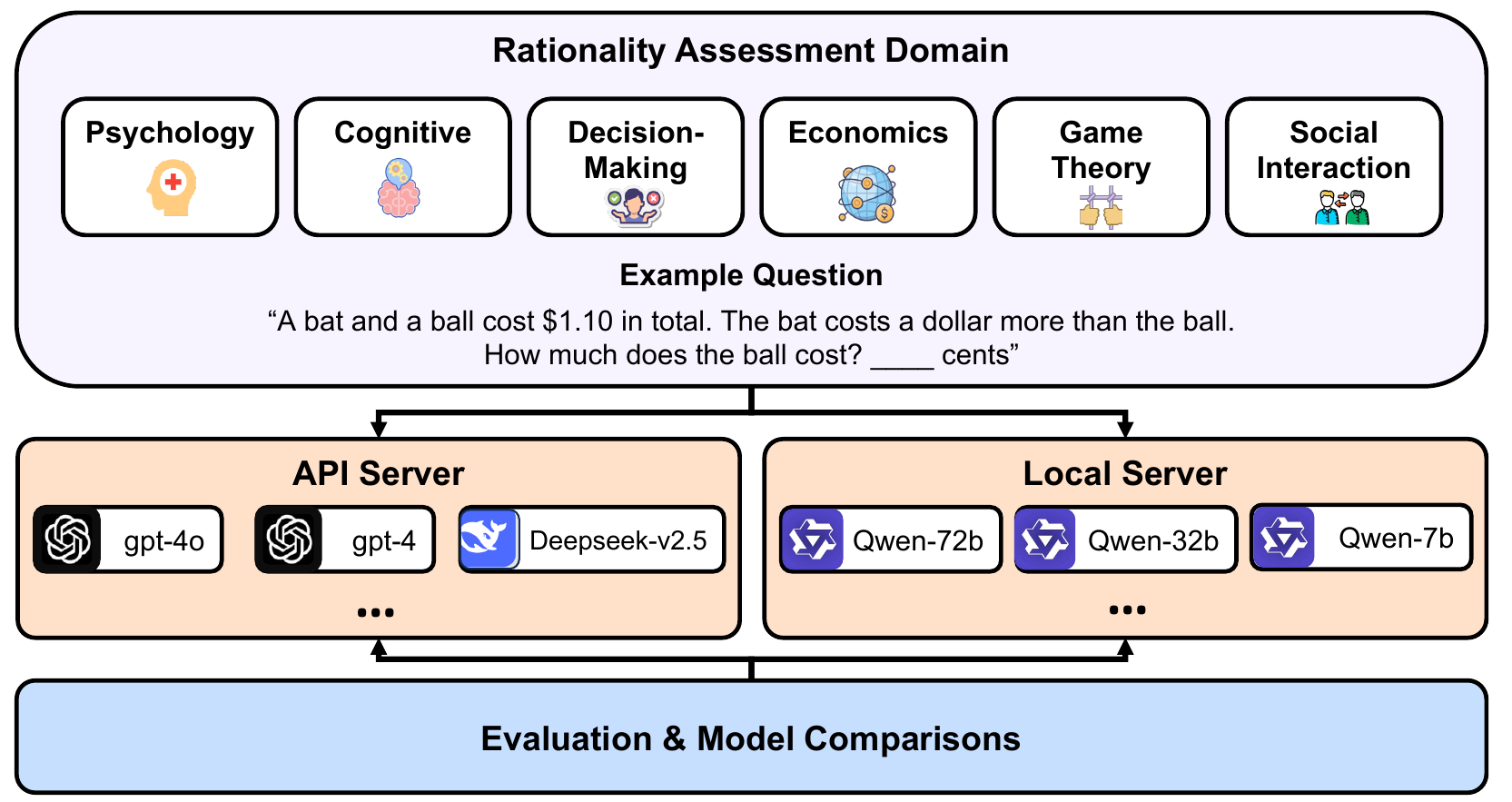}}
    \caption{Benchmark for assessing the rationality of LLMs across multiple domains.}
    \vspace{-10px}
    \label{fig:bench}
\end{figure}

The illustration of the benchmark is shown in Figure~\ref{fig:bench}. Under this framework, the user can specify the domain, aspect, or LLM. The benchmark toolkit can call and receive responses from both open-source and commercial API-based models, calculate the quantitative results of rationality, and provide an analysis of the human-LLM comparison. In the following, we present heatmap results (Figures~\ref{fig:psy}-~\ref{fig:social}), normalized by human performance, comparing different LLM models. Higher rationality is shown in orange (better than human performance), while blue indicates lower rationality than humans. Human performance data is sourced from literature reviews, where specific scales are measured by the LLMs. We also conduct further analysis on cross-domain correlations, impact of model size, data contamination, and comparison with reasoning models in Appendix~\ref{sec:supple_results}.
We present the original rationality scores in Appendix~\ref{app:original_rationality_scores}


\subsection{Psychology Domain}
\begin{figure}[h!]
    \centering
    \subfloat[Theoretical rationality]{\includegraphics[width=0.98\linewidth]{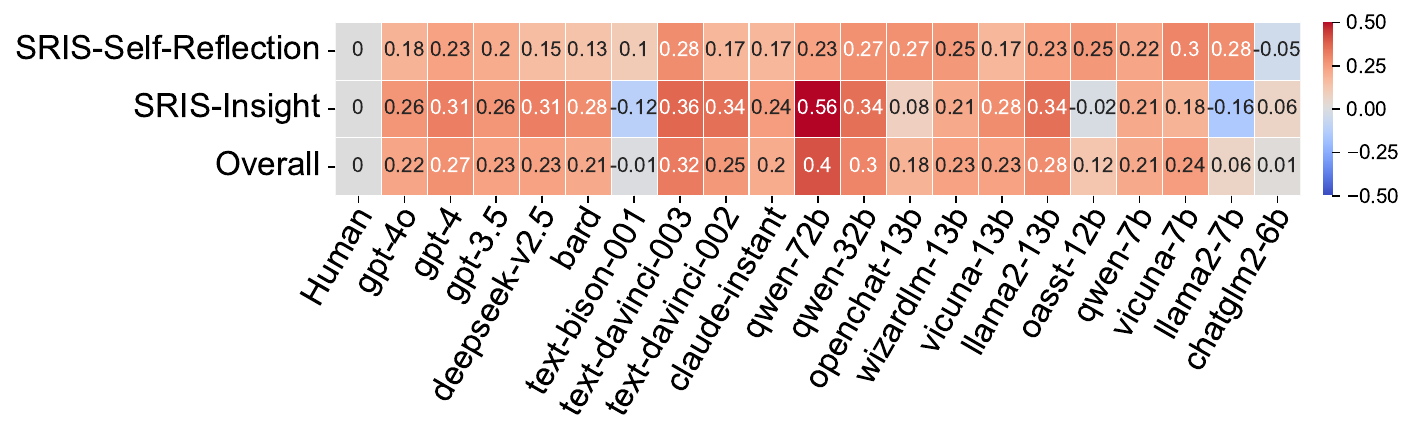}} \ \vspace{0.1cm}
      \subfloat[Practical rationality]{\includegraphics[width=0.97\linewidth]{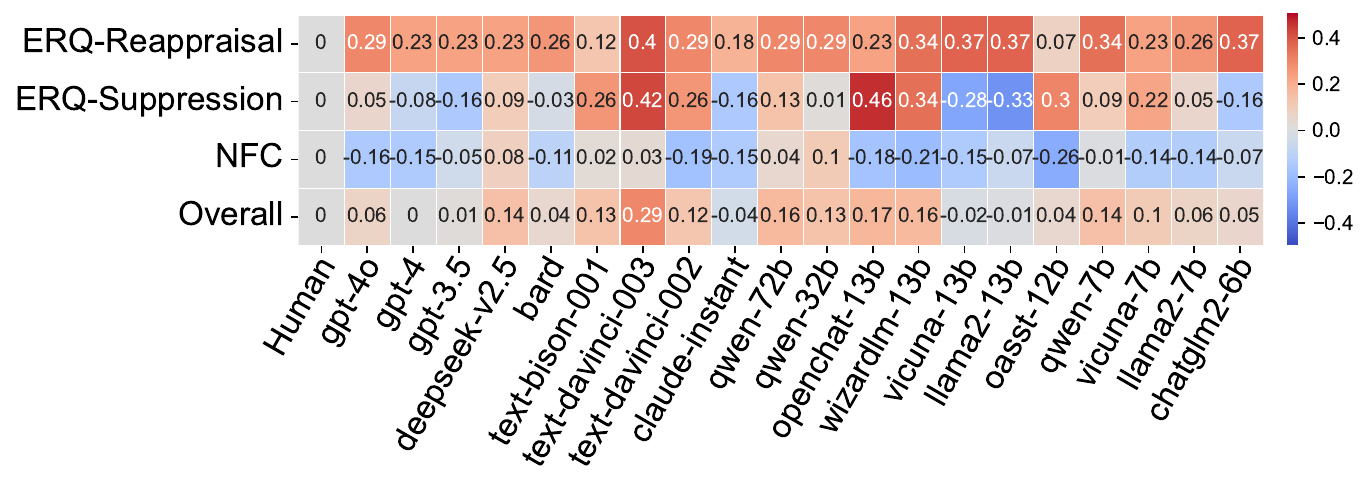}} \ \vspace{0.1cm}
     \vspace{-10px}
    \caption{Normalized rationality scores of LLMs (x) on different aspects (y) in psychology domain.}
    \vspace{-5px}
    \label{fig:psy}
\end{figure}

In the psychology domain, LLMs appear to be better than an average human in self-reflection and emotional reappraisal and worse than human in emotional suppression and motivation towards higher cognitive effort tasks.
As shown in Figure~\ref{fig:psy}a, most LLMs, except the older models, e.g., chatglm2-6b and text-bison-001, are more capable of introspection and understanding their own thoughts than average human. All LLMs outperform humans at emotional reappraisal, which means they are better at cognitively reframing emotional situations. However, older LLM models perform better at emotion suppression, while advanced models struggle with it. For NFC, DeepSeek and Qwen-72b, Qwen-32b slightly outperform humans, indicating they have greater motivation for engaging more complex cognitive tasks. Models, such as GPT-4o and GPT-4 are less inclined to engage in higher cognitive load tasks.
as illustrated in Figure~\ref{fig:psy}b.

\subsection{Cognitive Science Domain}
\begin{figure}[h]
    \centering
    \subfloat[Theoretical rationality]{\includegraphics[width=0.98\linewidth]{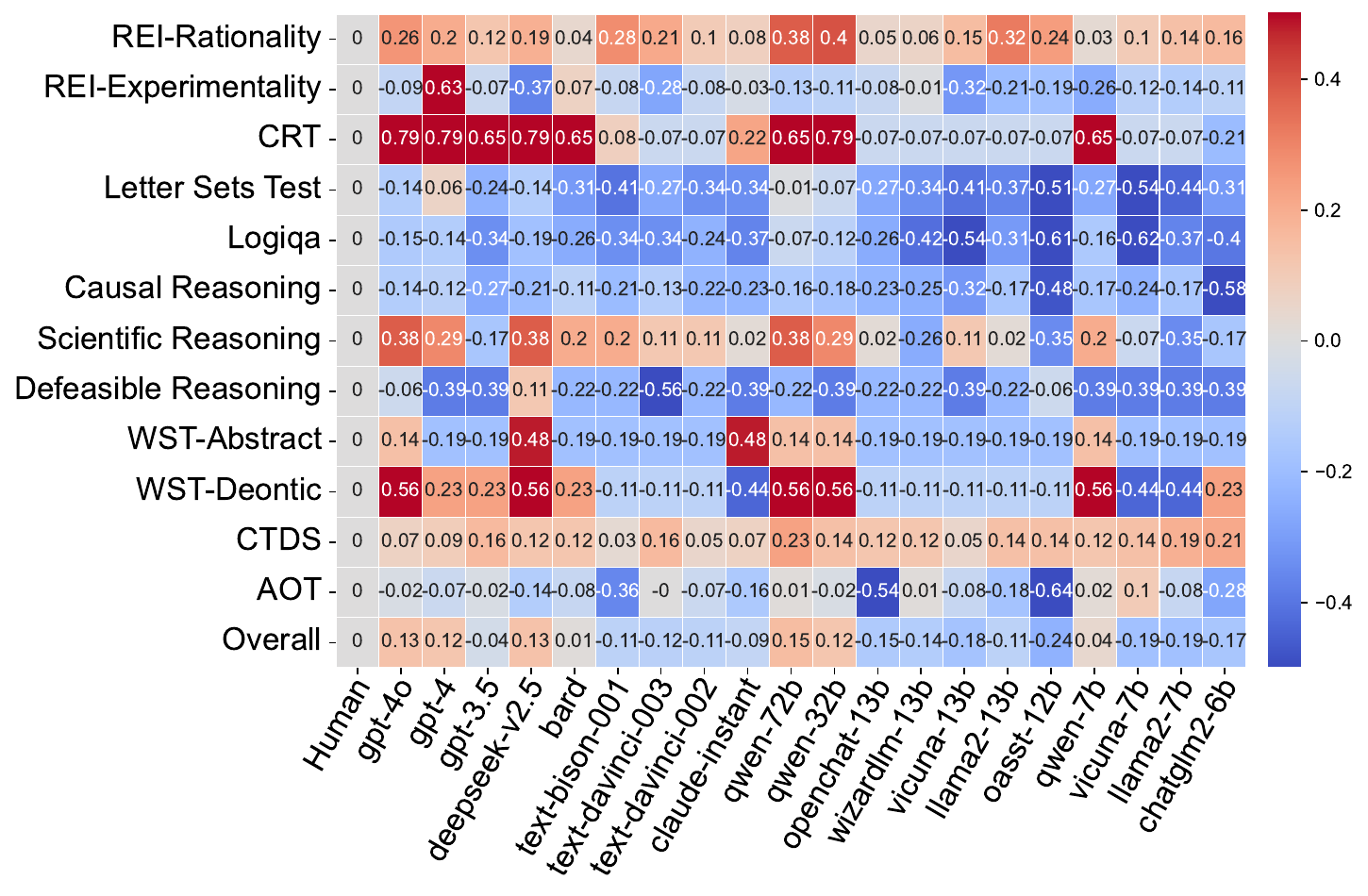}} \ \vspace{0.1cm}
      \subfloat[Practical rationality]{\includegraphics[width=0.97\linewidth]{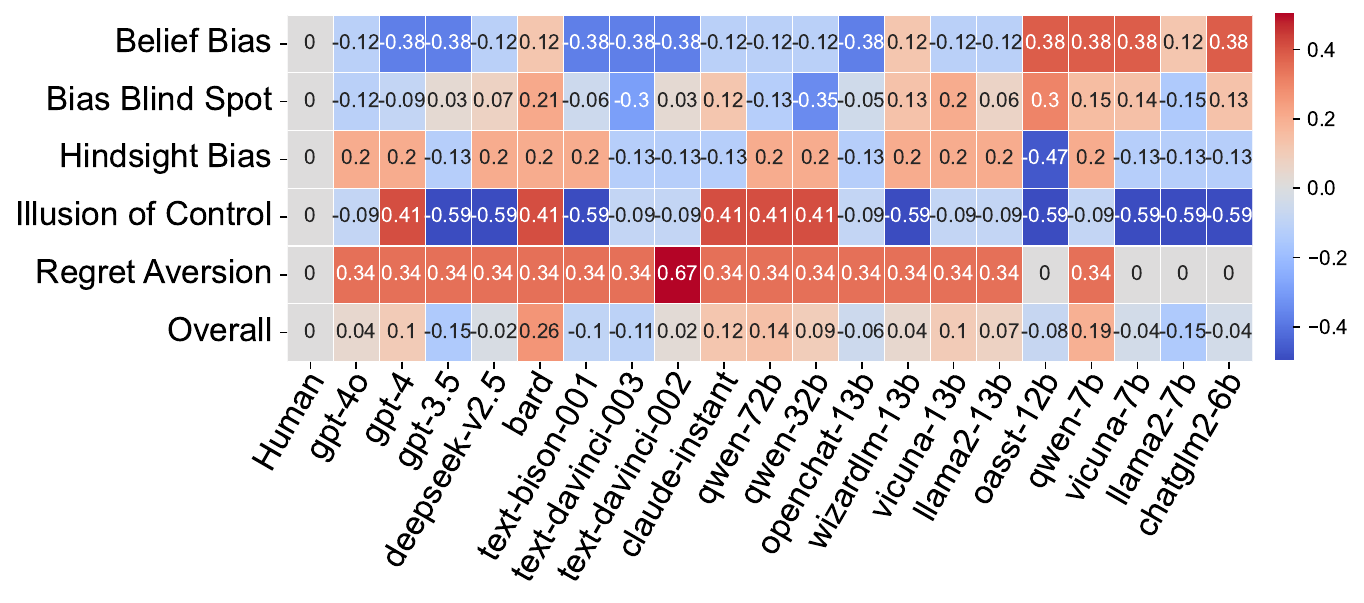}} \ \vspace{0.1cm}
    \vspace{-5px}
    \caption{Normalized rationality scores of LLMs (x) on different aspects (y) in the cognitive science domain.}
    \vspace{-5px}
    \label{fig:cog}
\end{figure}

In the cognitive science domain, as depicted in Figure~\ref{fig:cog}, advanced LLMs such as GPT-4o, GPT-4, DeepSeek v2.5, Qwen-7b, Qwen-32b, Qwen-72b, and Bard outperform humans in several reasoning tasks, including rational thinking, scientific reasoning, abstract reasoning, deontic reasoning, and critical thinking dispositions. These models are also less susceptible to cognitive biases such as regret aversion, consistently opting for the most optimal decision, regardless of potential regret or hindsight bias, indicating a stronger focus on rational decision-making.

As shown in Figure~\ref{fig:cog}a, LLMs consistently prefer rational thinking over intuitive approaches as revealed by REI. Models like Qwen-72b, Qwen-32b, and GPT-4o show a significantly stronger preference for rationality compared to humans. While all models (except GPT-4) also rated as more inclined to engage in intuitive thinking than humans. In terms of performance on the CRT, advanced LLMs all performed well and well than human. In general, results from CRT, REI, CTDS, and AOT indicate that advanced LLMs prefer rational thinking more than other models and humans. Moreover, DeepSeek v2.5 stands out in defeasible reasoning, a task that is particularly difficult for humans. However, in tasks like logical reasoning, including inductive, deductive, and causal reasoning, humans outperform LLMs. This is surprising and suggests that future research could investigate the varying levels of difficulty in these reasoning tasks. In general, as we move from older to more advanced LLMs, there is a clear improvement in their performance on cognitive science tasks. Finally, regarding the tasks involving cognitive biases, as shown in Figure~\ref{fig:cog}b, LLMs generally perform better than humans, particularly in regret aversion, where almost all models (except smaller ones) outperform humans. Notably, Bard excels across all cognitive bias tasks compared to humans. However, advanced models like DeepSeek and GPT-4o struggle with biases such as illusion of control, bias blind spot, and belief bias, indicating that while these models excel in theoretical rationality, they still face challenges in applying rational decision-making in practical scenarios.

\subsection{Decision-making Domain}

\begin{figure}[h!]
    \centering\includegraphics[width=0.98\linewidth]{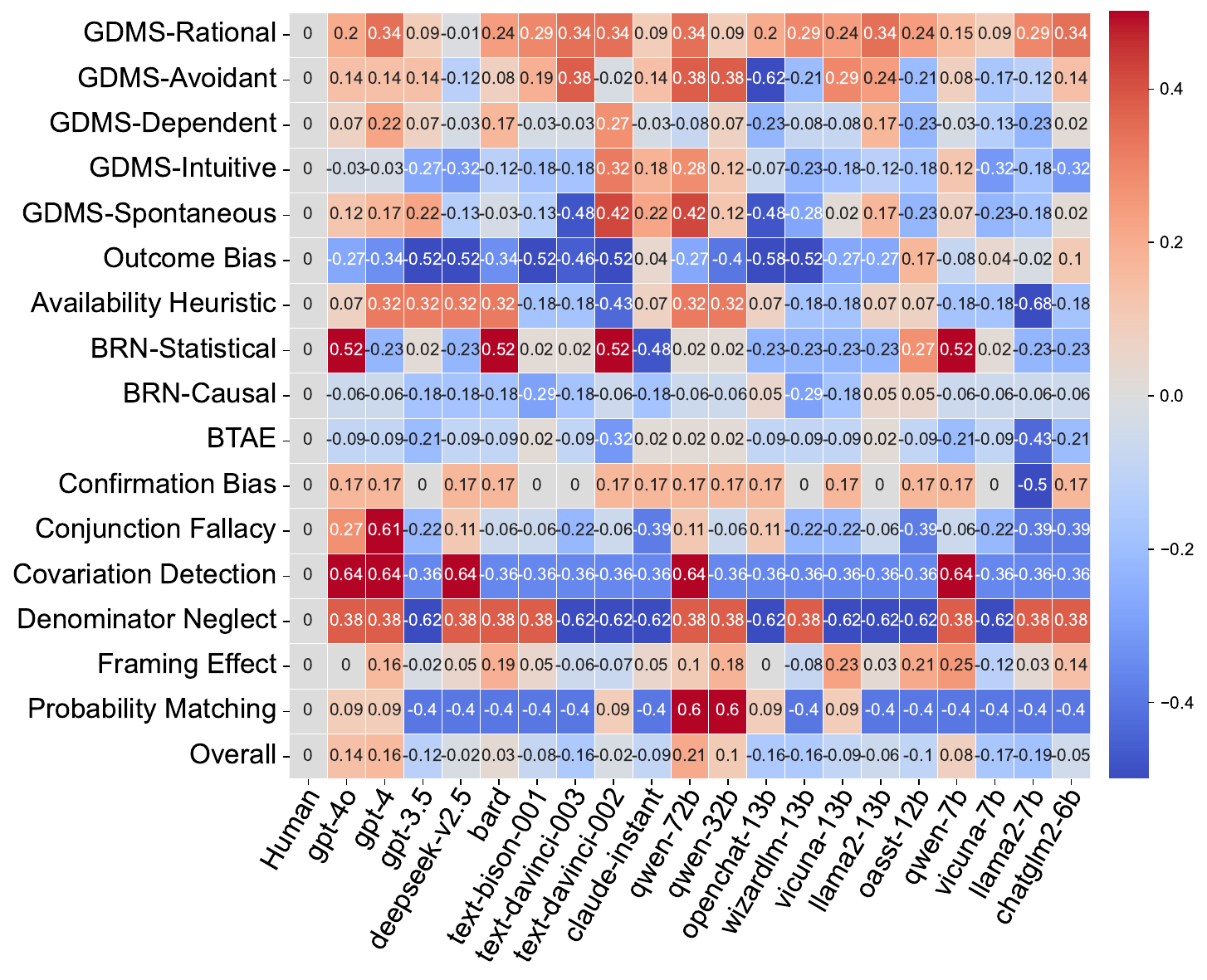}
     \vspace{-10px}
    \caption{Normalized rationality scores of LLMs (x) on different aspects (y) in decision-making domain.}
    \vspace{-5px}
    \label{fig:dec}
\end{figure}

In the decision-making domain, as shown in Figure~\ref{fig:dec}, all measured LLM models rate themselves as more rational than humans. Most models, especially newer ones, identify as avoidant, dependent, and spontaneous decision-makers, but are less intuitive than humans, with the exception of the Qwen models. Furthermore, GPT-4o, GPT-4, DeepSeek v2.5, Qwen-72b, and Qwen-32b show significantly lower susceptibility to most decision-making biases, except for the Better-than-Average Effect and Causal Base-Rate Neglect biases.

\subsection{Economy Domain}

\begin{figure}[h!]
    \centering
    \includegraphics[width=0.98\linewidth]{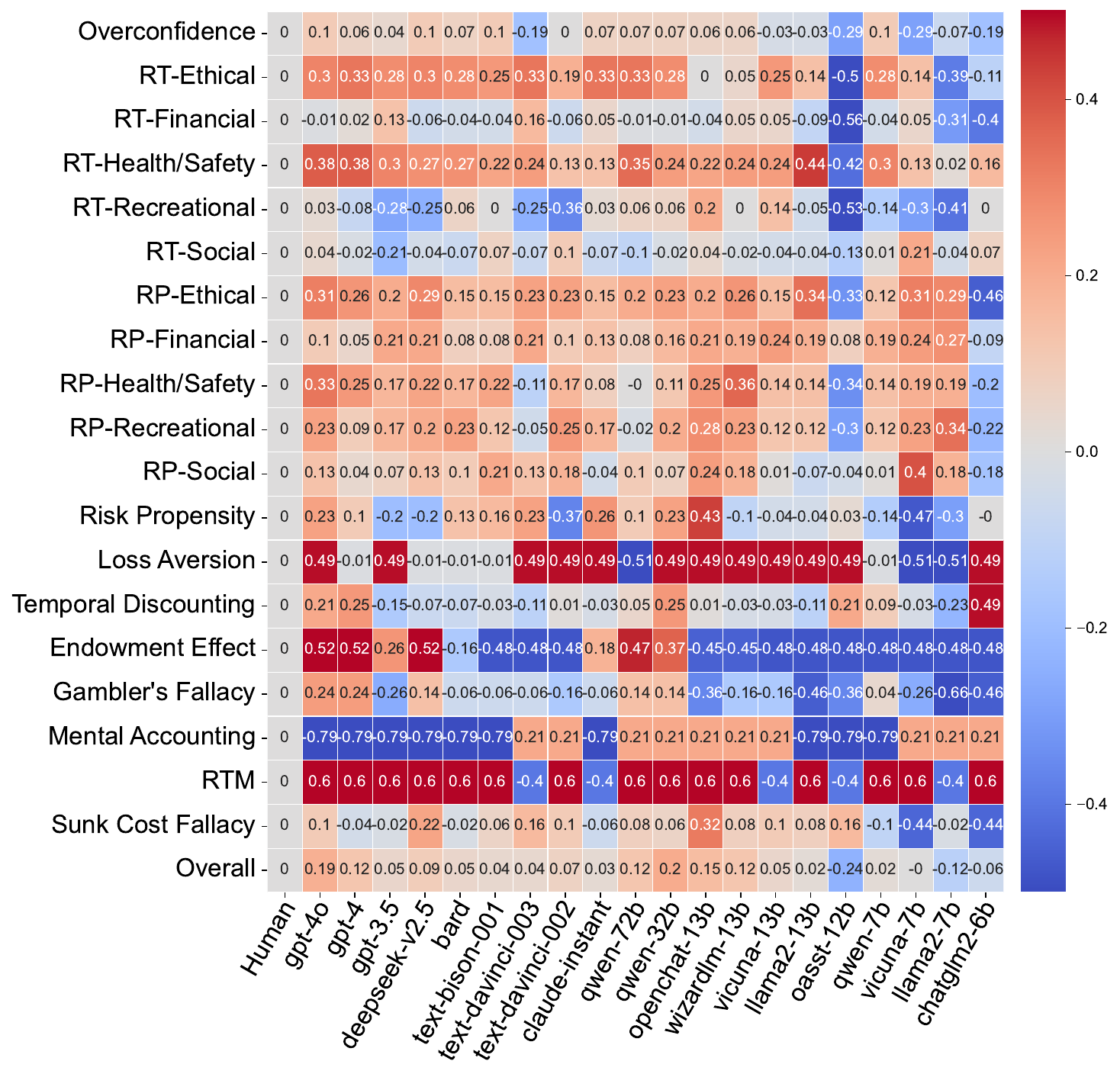}
     \vspace{-10px}
    \caption{Normalized rationality scores of LLMs (x) on different aspects (y) in economics domain.}
    \vspace{-7px}
    \label{fig:econ}
\end{figure}

The evaluation results in the economics domain are shown in Figure~\ref{fig:econ}. Risk-taking (RT) and risk perceptions (RP) are both measured across five different fields, such as financial, health, and recreational (also known as entertainment). We found that LLMs are overly concerned with risks relating to morality and health safety, far beyond the level of human concern. Interestingly, while LLMs are extremely concern about any type of risks, even entertainment risks, such as extreme sports, in action (i.e., for risk-taking), they often do not show the same level of concern as humans do for these potential risks.

\subsection{Game Theory Domain}
\begin{figure}[h!]
    \centering
    \includegraphics[width=0.98\linewidth]{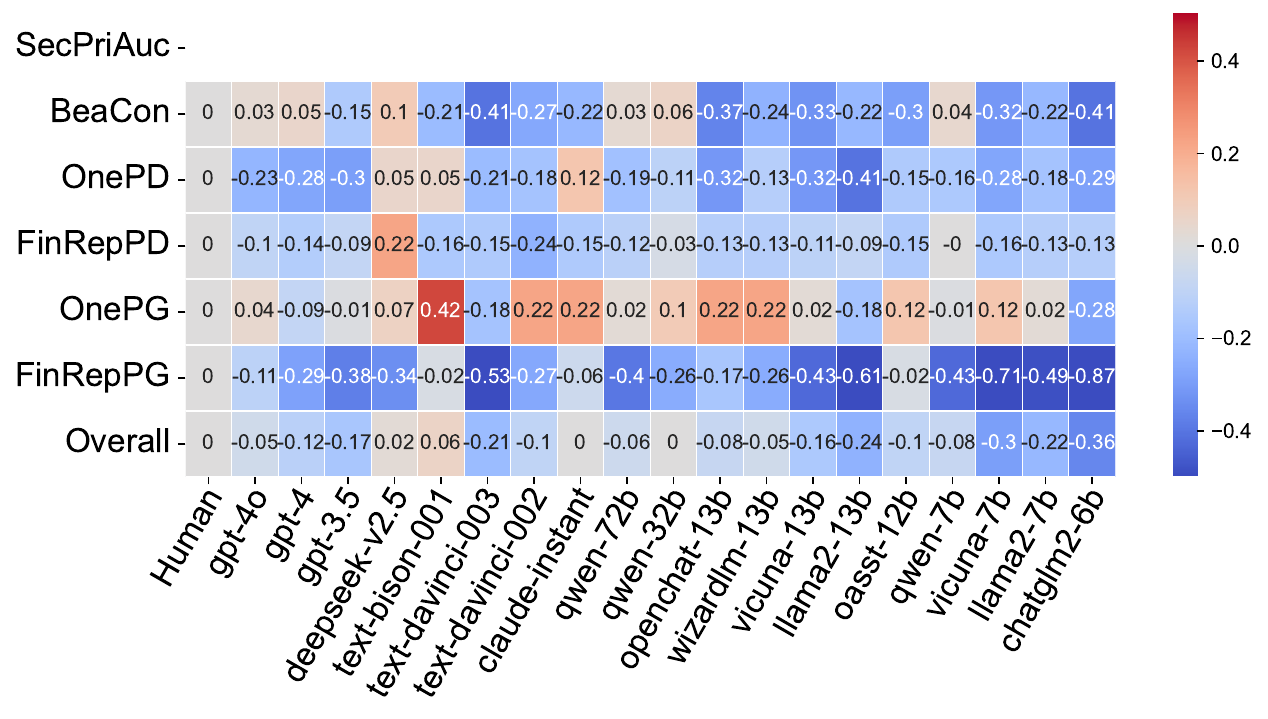}
     \vspace{-10px}
    \caption{Normalized rationality scores of LLMs (x) on different aspects (y) in game theory domain. (Missing values in the first row is due to the lack of human experiment data of the Second Price Auction game.)}
    \vspace{-10px}
    \label{fig:game}
\end{figure}

The evaluation results in the game theory domain are shown in Figure~\ref{fig:game}, of which the main conclusions are as follows.
Most of LLMs have lower rationality than humans in the game theory domain, including models that perform best in other domains like GPT-4. Only one LLM (text-bison) model has a higher overall rationality than humans. This phenomenon also holds for each specific game, where LLMs seldom surpass human performance. This result indicates that games are too complex for most LLMs to handle, where players need to anticipate others' strategies and respond accordingly.

The level of rationality is positively correlated with the number of parameters. The overall rationalities (average of all the games) of proprietary LLM models are generally higher than open-source models, and 13B models are also generally more rational than 7B models. Among them, text-bison achieves the highest overall rationality 0.78 in the game theory domain, surpassing human rationality 0.72. On the contrary, the smallest model chatglm2-6b exhibits the lowest rationality of 0.36.

Fine-tuning increases the level of rationality. We note that openchat-13b, vicuna-13b, and wizardlm-13b are fine-tuned from Llama model, and they exhibit higher overall rationality than llama-13b, indicating that the fine-tuning process may have a positive effect on LLM's rationality in game theory.

\subsection{Collective Rationality Domain}
\begin{figure}[h!]
    \centering
    \subfloat[Cooperation and coordination]{\includegraphics[width=\linewidth]{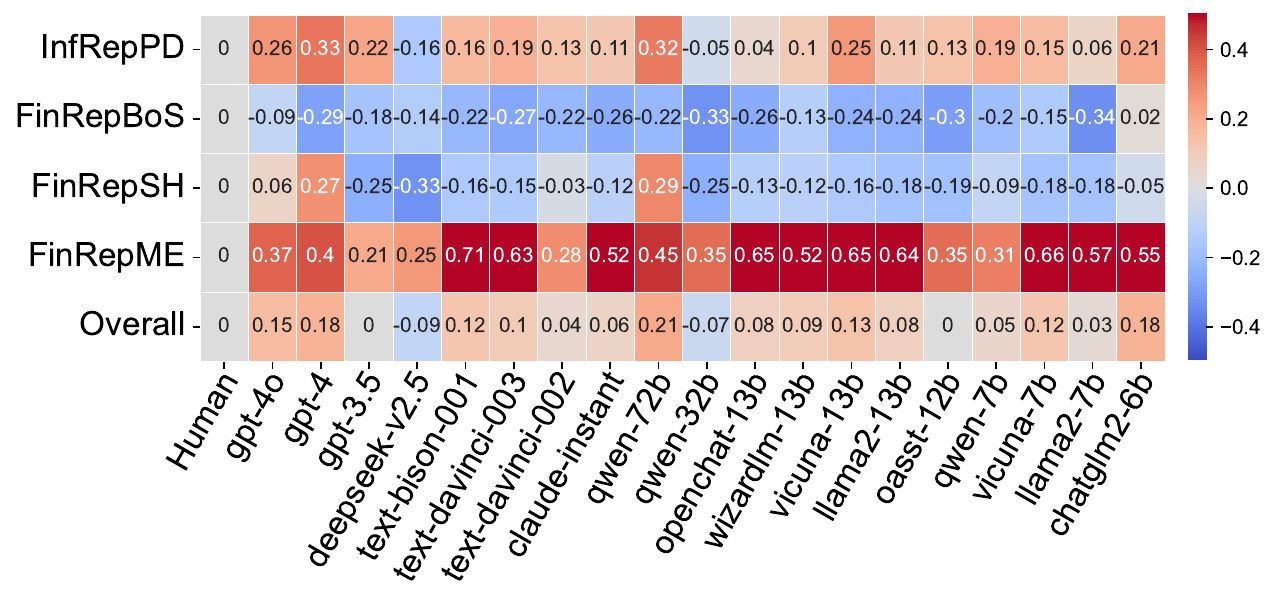}} \ \vspace{0.02cm}
      \subfloat[Wisdom of crowds - MMLU and MATH]{\includegraphics[width=\linewidth]{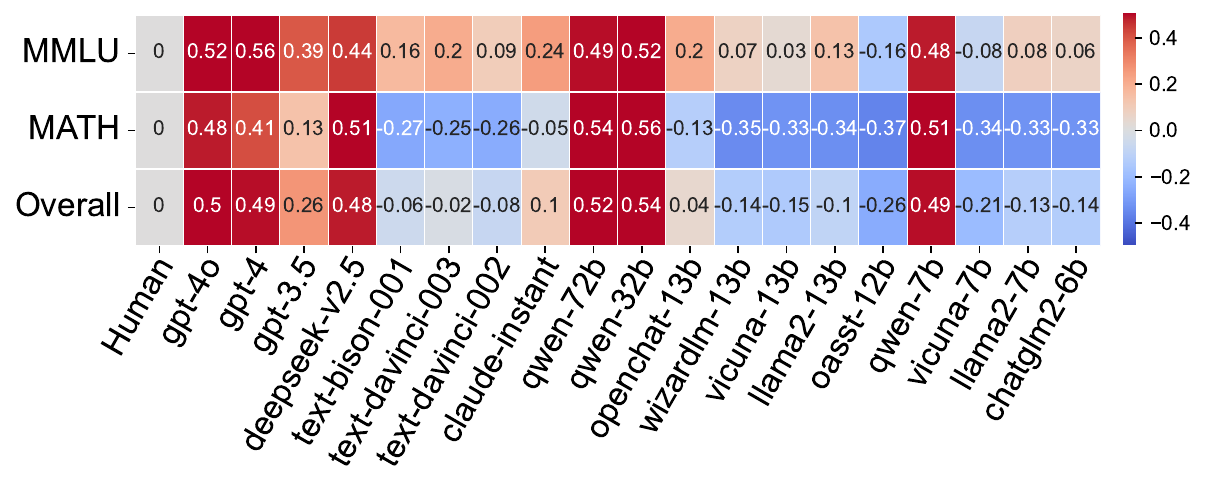}} \ \vspace{0.02cm}
     \subfloat[Wisdom of crowds - General knowledge]{\includegraphics[width=0.99\linewidth]{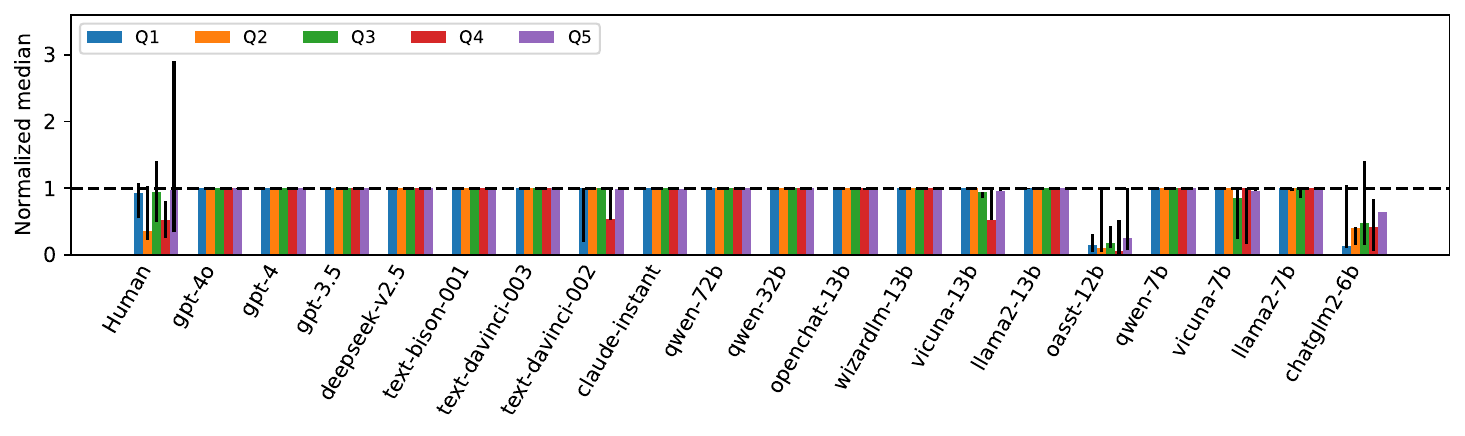}} \ \vspace{0.02cm}
    \caption{Normalized rationality scores of LLMs (x) on different aspects (y) in collective rationality domain.}
    \label{fig:social}
    \vspace{-10px}
\end{figure}

The evaluation results in the collective rationality domain are shown in Figure~\ref{fig:social}, of which the main conclusions are as follows.

LLMs generally perform better than humans in cooperation and coordination games. As shown in Figure~\ref{fig:social}(a), the overall rationality scores of almost all LLMs are higher than that of humans. This is probably because LLMs have relatively lower variance and can maintain stable strategies. For example, in minimum-effort games, LLMs tend to consistently choose values that lead to high efficiency, resulting in much higher rationality scores than humans. On the contrary, in a battle of the sexes where the best strategy is to alternate between two actions, most LLMs fail to establish such coordination with the other player.

Larger models demonstrate higher rationality in the wisdom of crowds experiments. As shown in Figure~\ref{fig:social}(b), gpt-4 and gpt-3.5 outperform humans by a considerable margin. In addition, the rationality score generally decreases as the model size becomes smaller (from left to right). Moreover, in the general knowledge test (shown in Figure~\ref{fig:social}(c), the closer to 1, the more rational), it can be observed that most LLMs correctly answer the five questions no matter how we set the agent profiles, while human answers vary greatly across different people.

\section{Conclusions}\label{sec::conclusion}
In this paper, we propose the first benchmark for evaluating the rationality of LLMs.
The results reveal insightful observations in different domains and various model characteristics, especially for the comparisons with humans.
The proposed toolkit of the benchmark can benefit the developers for training better LLMs, especially for AI-human alignment, and the users for assessing whether the specific LLM is capable and suitable for a given application.

\section*{Limitations}

\textbf{Different Characteristics in Measurement and Assessment Between LLMs and Humans:} Due to human-AI alignment regulations and ethical concerns, LLMs may be restricted from responding to certain questions, which can hinder direct feedback. Unlike humans, LLMs can operate continuously without fatigue, ensuring consistent processing, whereas human respondents may experience fatigue, affecting the quality of their responses. Additionally, humans might manipulate their responses or show inconsistencies, leading to inaccuracies. Detecting and addressing these discrepancies is crucial for robust evaluation. Our evaluation covers a selection of LLMs and may not represent all existing models or relevant criteria, so broader evaluation approaches are needed.

\textbf{Data Contamination in Rationality Evaluation:} LLMs may have encountered similar rationality questionnaires during training, leading to overestimation of their rationality by recalling answers rather than reasoning. This data contamination is a recognized issue, and methods for detection and mitigation are being explored. Innovative approaches, such as psychometric techniques and dynamic synthesis of evaluation benchmarks, are needed to avoid data contamination.

\clearpage
\bibliography{bibliography}

\clearpage
\appendix

\section*{Appendix} \label{sec::appendix_measurement}

\section{Detailed Measurement and Evaluation}
\label{app:measurement_evaluation}

\subsection{Psychology}
In the field of psychology, we employ psychometric evaluations, commonly applied to human participants, to assess rationality from both theoretical and practical perspectives. Specifically, we measure self-reflection as a key element of theoretical rationality as well as emotional regulation and intrinsic motivation as vital aspects of practical rationality. The following are the detailed descriptions of each measured aspect.

\textbf{Self-reflection.} It is an introspective aspect that evaluates an individual's awareness and clarity of their thoughts, emotions, and behaviors~\cite{grant2002self}. It serves as a basis for the understanding and connection of arguments and judgments, leading to clearer and more logical thinking~\cite{bender2016reflection}. It also plays a role in debiasing interventions by bringing unconscious biases into conscious awareness~\cite{domeier2018motivational}, contributing to the enhancement of rationality. We measure LLMs' ability in self-reflection by adapting the Self-Reflection Insight Scale (SRIS) from Grant \textit{et al.}~\cite{grant2002self}. The scale is a 20-item self-rating questionnaire that consists of two major components: self-reflection (the introspection and self-evaluation of their thoughts, feelings, and behaviors) and insight (the clarity in acknowledging their inner self). All our used questionnaires are available on GitHub. Several questions are reverse-scored (i.e., the score on the item indicates the opposite of the intended aspect), and a high average score on this scale is evaluated as being more rational. 
    
\textbf{Emotion regulation.} It is the ability to control, evaluate, and adjust one's emotional state~\cite{garnefski2007cognitive}. Emotion can provoke errors or even hinder goal achievement if the individual neither intends nor decides to adopt the emotional state~\cite{lambie2008irrationality}. Prior literature suggests that effective emotion management can indicate emotional rationality~\cite{kerr2021rationality}. Wang \textit{et al.}~\cite{wang2023emotional} illustrate that many LLMs, such as GPT 4, GPT 3.5 turbo, ChatGLM, and text-davinci-001, are capable of understanding emotions and have performed better than average humans on an emotional intelligence test. In our study, we assess LLMs' emotion regulation ability using the Emotion Regulation Questionnaire (ERQ), which was a 10-item scale employed by Gross \textit{et al.}~\cite{gross2003individual} on human subjects. It consists of two aspects of emotional regulation: cognitive reappraisal and expressive suppression. Cognitive reappraisal represents reevaluating situations to ameliorate their emotional impact, while expressive suppression refers to the deliberate control of one's emotions~\cite{gross2003individual}. A higher rating on cognitive reappraisal and expressive suppression is scored as being more rational.

\textbf{Intrinsic motivation.} It refers to an individual's internal drive to undertake challenging cognitive tasks~\cite{cacioppo1984efficient}. When people are intrinsically motivated, they are more inclined to engage in rational thinking and decision-making processes~\cite{curcseu2012decision,knauff2021handbook}. In this study, we used the 18-item Need for Cognition (NFC) scale developed by Cacioppo \textit{et al.}~\cite{cacioppo1984efficient} to assess individuals' willingness to engage in active cognitive activities. As indicated in prior literature, NFC is positively correlated with rationality in decision-making processes~\cite{curseu2006need}. For the scale evaluation, several questions are reverse-scored and a high overall average score is representative of greater rationality.

In general, for the aspects mentioned above, we instruct each LLM to act as a participant to answer the questionnaires. Afterward, we calculate their average score and normalize it to a rationality score ranging from 0 to 1, with 0 representing the least rational and 1 representing the most rational.

An example prompt from the psychology domain is shown below.
\begin{tcolorbox}[breakable, enhanced, colback=gray!10, colframe=black, boxrule=0.4pt]
\small
\textbf{Prompt for Self-reflection:}\\
Please act like a participant in this survey. For the following statements, please respond to each statement by selecting a number from 1 to 6. This number should best represent your opinion on a 6-point Likert scale (1 = strongly disagree, 2 = disagree, 3 = somewhat disagree, 4 = somewhat agree, 5 = agree, 6 = strongly agree).\\
1. I don’t often think about my thoughts\\
2. I rarely spend time in self-reflection\\
3. I frequently examine my feelings\\
4. I don’t really think about why I behave in the way that I do\\
5. I frequently take time to reflect on my thoughts\\
6. I often think about the way I feel about things\\
7. I am not really interested in analyzing my behavior\\
8. It is important to me to evaluate the things that I do\\
9. I am very interested in examining what I think about\\
10. It is important to me to try to understand what my feelings mean\\
11. I have a definite need to understand the way that my mind works\\
12. It is important to me to be able to understand how my thoughts arise \\
\end{tcolorbox}

\subsection{Cognitive Science}
In the field of cognitive science, similar to psychology, we evaluate LLMs' rationality from both theoretical and practical approaches. For theoretical rationality, we assess their cognitive processes (i.e., dual process theory), logical reasoning (i.e., inductive, deductive, and causal reasoning), context-based reasoning (i.e., defeasible reasoning, scientific reasoning, and deontic reasoning), and thinking dispositions (i.e., critical thinking and open-minded thinking). Understanding LLMs' cognitive processing styles can offer insights into the underlying mechanisms that govern their behaviors~\cite{stanovich2000individual,evans2003two}. Logical reasoning is the fundamental capability of rationality~\cite{arthur1991designing,weber2014logic}. 
These various forms of context-specific reasoning are essential for making conditional and situational judgments, as they adapt to the unique aspects of each situation, allowing for a more tailored approach to the rationality assessment~\cite{knauff2021handbook,ragni2017wason}. Thinking dispositions. including critical as well as open-minded thinking dispositions represent individuals' cognitive styles of thinking and they substantially relate to rationality~\cite{stanovich2000individual,erceg2022normative}. Lastly, we assess practical rationality through a range of cognitive biases. As discussed in numerous studies, the use of biases often contributes to systematic errors or irrational thinking~\cite{kahneman2003maps,erceg2022normative,kahneman2003maps}. In our study, we engage LLMs as participants, prompting them to respond to a variety of questionnaires that measure different aspects related to rationality. Below are the details of the facets of rationality that we measure.

\textbf{Dual process theory.} This theory is often described as the architecture of cognition that differentiates between two cognitive modalities: System 1 (i.e., intuitive, rapid, and automatic thinking) and System 2 (i.e., rational, analytic, and controlled thinking)~\cite{knauff2021handbook,evans2003two,de2008conflict}. In our study, we assess these processes through two scales: Rationality-Experimental Inventory (REI) and Cognitive Reflection Test (CRT). Specifically, we adapt the REI scale from Pacini and Epstein~\cite{pacini1999relation}, which measures an individual's tendency to engage in rational and experimental thinking processes. A rational thinking style is characterized by a tendency to make judgments analytically and logically (``System 2''), while an experimental thinking style is associated with the inclination to make decisions based on intuitions, feelings, and immediate thoughts (``System 1'')~\cite{phillips2020rational}. Note that both rationality and experimentality can be further broken down into two subdimensions, namely ability and engagement; however, for our study, we consider them collectively. In our evaluation, a higher average score on the rational component of the scale and a lower average score on the experimental component are both considered as indications of being more rational. Moreover, we measure CRT as the tendency to refrain from intuitive (``System 1'') yet incorrect responses in favor of reflective thinking (``System 2''). Erceg \textit{et al.}~\cite{erceg2022normative} have found that CRT is highly correlated with rationality. In our study, we utilize the 7-item CRT proposed by Toplak \textit{et al.}~\cite{toplak2014assessing} and calculate the correct rate as a measurement of the degree of rationality (e.g., a higher correction rate indicating a greater degree of rationality).

\textbf{Inductive reasoning.} As a major component of logical reasoning, inductive reasoning refers to the ability to derive general conclusions from specific observations or premises~\cite{goswami2010inductive}. We utilize the Letter Sets, combining part 1 and 2 of the test in the Kit of Factor-Referenced Cognitive Test developed by Ekstrom \textit{et al.}~\cite{ekstrom1976kit}, to assess inductive reasoning. Our test consists of 30 items, with each question presenting five groups of four letters, and four of the sets follow the same rule, while one does not. Participants are tasked to identify the set with different rules. The rationality degree is measured by the accuracy of the answers.

\textbf{Deductive reasoning.} As another key aspect of logical reasoning, deductive reasoning involves the ability to apply general rules to obtain specific conclusions~\cite{goswami2010inductive}. Adapting from Liu \textit{et al.}~\cite{liu2023logiqa}, we randomly selected 100 multiple-choice questions from the ``test.txt'' dataset provided by the researchers on their GitHub to evaluate the LLMs. This dataset covers five subcategories of deductive reasoning: categorical, sufficient conditional, necessary conditional, disjunctive, and conjunctive reasoning. LLMs are assessed on their ability to categorize aspects, interpret ``if P, then Q'' conditional statements, understand necessary conditions (e.g., ``P only if Q''), handle scenarios with two premises where only one is needed for the conclusion, and determine when the conclusion holds true if and only if all premises are true. LLMs' rationality scores are determined based on their overall accuracy in responding to all 100 questions.

\textbf{Causal reasoning.} Causal reasoning, critical for understanding cause-and-effect relationships and their underlying logic, often intersects with both inductive and deductive reasoning, playing a core role in our interpretation of the world~\cite{eells2016rational,binz2023using}. To measure causal reasoning, we employ a set of questions originally developed by Waldman and Hagmayer~\cite{waldmann2005seeing} and then modified by Binz and Schulz~\cite{binz2023using}. The questionnaire composites both common-cause scenarios (e.g., a shared causal variable ``A'' causes two other variables ``B'' and ``C'') and causal-chain scenarios (e.g., each variable sequentially causes the other: ``B'' causes ``A'', and ``A'' then causes ``C'' ). In particular, interventions are utilized in the test to better scrutinize the inherent causal systems of LLMs by manipulating a variable while observing the consequences of other variables. Each LLM receives a rationality score based on the absolute difference between its responses and the ideal answers.

\textbf{Defeasible reasoning.} This is characterized by the ability to withdraw or revise previously held beliefs or background knowledge in response to new information. This form of reasoning is adaptable, allowing for adjustments based on emerging evidence or changing situations~\cite{chen2023jailbreaker,ford2000strategies}. We apply the six defeasible reasoning questions designed by Ford and Billington~\cite{ford2000strategies} to assess LLMs' abilities. Rationality score is based on the accuracy of their answers.

\textbf{Scientific reasoning.} Scientific reasoning, rooted in scientific methods and principles, evaluates LLMs' competence in handling scientific evidence, contributing to our understanding of LLMs' scientific rationality~\cite{knauff2021handbook, drummond2017development}. Recognizing the accuracy and reliability of these models within research contexts is crucial for determining their appropriate application in research and acknowledging their limitations as research assistants, as prior researchers have pointed out~\cite{liu2023trustworthy}. It guides us in effectively leveraging LLMs in scientific inquiries and in making informed decisions about their use. We employ the scientific reasoning scale developed by Drummond and Fischhoff~\cite{drummond2017development}, which is applicable even for individuals with no scientific background to evaluate their scientific reasoning skills. The rationality score is determined based on the correctness rate of LLMs' responses.

\textbf{Deontic reasoning.} This involves assessing whether actions follow or violate the normative rules and duties, and thus practice social and moral reasoning~\cite{ragni2017wason}. To assess the reasoning, we include two sets of Wason Selection Task questions, a total of six questions, developed by Ragni \textit{et al.}~\cite{ragni2017wason} and Erceg \textit{et al.}~\cite{erceg2022normative}. Half of the questions utilize abstract conditional reasoning, and the other half evaluate deontic reasoning. LLMs' abilities in both reasoning types are evaluated. The correctness rate of LLMs' responses, separated by the two reasoning types, determines the rationality score.

\textbf{Critical thinking disposition.} This refers to the inclination to analytically and logically think and evaluate information or situations~\cite{sosu2013development,betz2020critical}. In our study, we applied the Critical Thinking Disposition Scale proposed by Sosu~\cite{sosu2013development}. The scale comprises two components: critical openness and reflective skepticism. Critical openness refers to the willingness to embrace new ideas, critically analyze them, and adjust beliefs based on new evidence. Reflective skepticism is the inclination to learn through reflecting on prior experiences and thoroughly evaluating evidence. The rationality score is determined by the overall average rating on the scale.
    
\textbf{Open-minded thinking disposition.} Open-minded thinking disposition refers to the tendency to consider diverse and alternative ideas, objectively assess evidence that contradicts one's beliefs, and reflect on current thoughts~\cite{stanovich2023actively}. In our study, we apply the Actively Open-Minded Thinking scale, developed by Campitelli and Gerrans~\cite{campitelli2014does}, to evaluate rationality. This approach is supported by prior studies as a measurement of rational thinking~\cite{baron2022actively,bensley2023critical,erceg2022normative}. The rationality score is determined by the overall average rating on the scale.
    
\textbf{Cognitive biases.} Cognitive biases are mental shortcuts often used to reduce cognitive load and facilitate rapid judgments~\cite{erceg2022normative,knauff2021handbook}. They are known as systematic errors in reasoning and evaluation~\cite{berthet2023heuristics}. In this case, our assessment of LLMs focuses on specific biases closely related to rationality~\cite{berthet2023heuristics}. These biases include belief bias in syllogistic reasoning, bias blind spot, hindsight bias, illusion of control, and regret aversion. \textbf{Belief bias in syllogistic reasoning} refers to the tendency to judge an argument's validity based on the believability of the conclusion, rather than on logical consistency~\cite{markovits1989belief}. We implement the measured scale from Berthet~\cite{berthet2021measurement}, originally developed from Teovanović \textit{et al.}~\cite{teovanovic2015individual}. \textbf{Bias blind spot} is the tendency to perceive oneself as less biased than others~\cite{scopelliti2015bias,berthet2023heuristics}. We apply the scale from Scopelliti \textit{et al.}~\cite{scopelliti2015bias}. \textbf{Hindsight bias} represents the tendency to view past events as more predictable than they actually were~\cite{teovanovic2015individual}. \textbf{Illusion of control} refers to the tendency to overestimate one's control of situations~\cite{rieger2022survey}. \textbf{Regret aversion} is the tendency to make decisions to avoid future regret~\cite{berthet2023heuristics}. We assess the performance of LLMs on these three biases, adopting tests from Rieger \textit{et al.}~\cite{rieger2022survey}, to evaluate their susceptibility. For all applied scales, we determine the presence of these biases in LLMs and inversely adjust the final scores to reflect rationality levels.
    
In sum, similar to the evaluation of psychology aspects, we compute an average score based on self-assessment or accuracy for the scales and normalize the final scores to reflect rationality levels, where 0 signifies the least rational and 1 signifies the most rational. 

An example prompt from the cognitive science domain is shown below.
\begin{tcolorbox}[breakable, enhanced, colback=gray!10, colframe=black, boxrule=0.4pt]
\small
\textbf{Prompt for Watson selection task:}\\
Please act like a participant in this survey. Imagine you are a police officer on duty. It is your job to ensure that people conform to certain rules. There are four cards shown to you that have information about four people sitting at a table. Each card is labeled with "Drinking beer", "Drinking coke", "22 years of age", and "16 years of age" on one side of the card, respectively. On one side of a card is a person's age and on the other side of the card is what a person is drinking. 
Here is a rule: If a person is drinking beer, then that person must be over 18 years of age. 
Select the cards that you need to turn over to determine whether or not the people are violating the rule.\\
a. the card labeled with "Drinking beer"\\
b. the card labeled with "Drinking coke"\\
c. the card labeled with "22 years of age"\\
d. the card labeled with "16 years of age"
\end{tcolorbox}

\subsection{Decision-making}
In the field of decision-making theory, research outlines representative patterns to categorize individuals by their \textbf{decision-making styles}. Scott and Bruce~\cite{scott1995decision} developed a self‐report questionnaire that delineates five decision-making styles: rational, intuitive, dependent, avoidant, and spontaneous. The rational style refers to the logical and structured approaches to decision-making; the intuitive style indicates the reliance upon hunches, feelings, and impressions; dependent style means the reliance upon the direction and support of others; the avoidant style entails postponing or avoiding decision-making; and spontaneous style is characterized by being impulsive and prone to making ``snap'' or ``spur of the moment''
decisions~\cite{scott1995decision,spicer2005examination}. We evaluate LLMs across different decision-making styles, interpreting a higher score in the rational style as a stronger propensity towards rationality and a lower score in all other styles as a weaker propensity towards rationality. 

Furthermore, we assess LLMs on various decision-making related heuristics and biases, the systematic deviations that are inherent in the decision-making processes: availability heuristics, base-rate neglect (statistical and causal), better-than-average, confirmation bias, conjunction fallacy, covariation detection, denominator neglect, framing effect, probabilistic matching, and outcome bias~\cite{berthet2023heuristics,ceschi2019dimensions}. \textbf{Availability heuristic} refers to the inclination to evaluate the likelihood or frequency of an event by how immediately examples come to one's mind. We assess the availability heuristic using the task from Erceg \textit{et al.}~\cite{erceg2022normative}, originally developed by Lichtenstein \textit{et al.}~\cite{lichtenstein1978judged}. \textbf{Base-rate neglect} is the propensity to neglect base-rate information in favor of specific instances. It has two forms: statistical base-rate neglect refers to general base-rate information, and causal base-rate neglect indicates causation-related base-rate information. We adopt the tasks provided by Erceg \textit{et al.}~\cite{erceg2022normative} to measure these aspects. \textbf{Better-than-average effect} describes the tendency for individuals to consider their abilities, traits, and characteristics to be above the average level of their peers. This bias is measured using the scale from Erceg \textit{et al.}~\cite{erceg2022normative}. \textbf{Confirmation bias} involves the tendency to interpret and favor information that aligns with one's beliefs and values. In this case, we assess the susceptibility of LLMs toward confirmation bias through a set of financial decision-making tasks~\cite{rieger2022survey}. \textbf{Conjunction fallacy} represents the tendency to erroneously believe that the combination of two events is more probable than either event occurring independently. We adapt the scenario task from Burgoyne \textit{et al.}~\cite{burgoyne2023understanding} to evaluate the bias. \textbf{Covariation detection} is the tendency to overestimate the relationship between two variables, often exacerbated by neglecting essential comparative (e.g., control group) information. We utilize the scenario task proposed by Toplak \textit{et al.}~\cite{toplak2014assessing} to measure the bias. \textbf{Denominator neglect}, also known as the ratio bias, indicates the tendency to focus excessively on numerators while neglecting denominators, affecting the judgment of probabilities, especially when presented in different ratio formats. We assess it using the task from Toplak \textit{et al.}~\cite{toplak2014assessing}. \textbf{Framing effect} refers to the tendency for people's decisions to be influenced by the way in which information is presented, particularly with respect to risk-choice and attribute framing. To evaluate both forms of this effect concurrently, we employ the task methodology from Bruine de Bruin \textit{et al.}~\cite{bruine2007individual}. \textbf{Probability matching} is the inclination of aligning the proportions of choices with the proportions of outcomes in a binary prediction task, rather than optimizing for the most likely outcome. The bias is measured by the dice problem and card guessing game implemented by West and Stanovich~\cite{west2003probability}. \textbf{Outcome bias} indicates the tendency to evaluate a decision based on the outcome rather than the quality of the decision at the time it was made. We employ Erecg \textit{et al.}~\cite{erceg2022normative} to measure the bias. In sum, a higher susceptibility to biases indicates a lower score in rationality.

An example prompt from the decision-making domain is shown below.
\begin{tcolorbox}[breakable, enhanced, colback=gray!10, colframe=black, boxrule=0.4pt]
\small
\textbf{Prompt for Base-rate neglect:}\\
Please act like a participant in this survey and answer the following questions.
Among the 1000 people that participated in the study, there were 995 nurses and 5 doctors. John is randomly chosen participant in this research. He is 34 years old. He lives in a nice house in a fancy neighborhood. He expresses himself nicely and is very interested in politics. He invests a lot of time in his career. Which is more likely?\\
  a. John is a nurse.\\
  b. John is a doctor.
\end{tcolorbox}



\subsection{Economics}

In the field of economics, rationality is commonly understood as the maximization of utility~\cite{kahneman2003maps}. It can also be categorized into theoretical and practical dimensions. Particularly, given the overlapping interests with other domains mentioned previously, we focus more on the unique aspects prominent in economics. In the theoretical aspect, \textbf{overconfidence} is a key factor influencing individuals' economic judgment. Overconfidence refers to individuals' inclination to overestimate their capabilities, which could lead to poor judgment. We measure the aspect using the metrics provided by Michailova \textit{et al.}~\cite{michailova2010development} where we compare the difference between LLMs' performance and their self-rated confidence level on the test. Rationality score is determined as the resistance to overconfidence.

For practical rationality, the evaluation contains several major aspects, including risk preference, time preference, and economic biases. \textbf{Risk preference} is measured by risk attitude, risk propensity, and loss aversion. In our study, we apply the revised version of domain-specific risk attitude scale developed by Blais and Weber~\cite{blais2006domain}. The scale measures across five different domains (i.e., ethical, financial, health or safety, recreational, and social) between two different facets (i.e., risk-taking and risk perception). Rationality degree is determined by the lower risk-taking and higher risk perception score towards these risk-related questions. Risk propensity, as another dimension of risk preference, is assessed using the 5-item risk propensity scale included in prior literature~\cite{bachmann2010risk}. Lower risk propensity indicates a higher degree of rationality. Furthermore, loss aversion, often considered as a variation in risk aversion~\cite{stango2019we}, is the tendency to prefer avoiding losses over acquiring  gains~\cite{chapman2017willingness}. To measure this aspect, we employ the task proposed by Stango and Zinman~\cite{stango2019we}, which captures the influence of loss aversion on decision-making processes. The degree of rationality is determined by the lower score of loss aversion. Moreover, \textbf{time preference} is assessed through a task that measures temporal discounting, which is the phenomenon of favoring more immediate rewards over future benefits. The task we utilize is adapted from Toplak \textit{et al.}\cite{toplak2014assessing}, originally developed by Frederick\cite{frederick2005cognitive}. Rationality is indicated by a lower propensity for temporal discounting.

Finally, we evaluate the susceptibility of LLMs to economic-related biases: endowment effect, gambler's fallacy, mental accounting, regression to the mean, and sunk cost fallacy~\cite{berthet2023heuristics,stango2019we}. \textbf{Endowment effect} is characterized by the tendency for individuals to value an object more once they own it, compared to when they do not. To measure this effect, we adapt the mug problem from Franciosi \textit{et al.}~\cite{franciosi1996experimental}, which was initially developed from Kahneman \textit{et al.}~\cite{kahneman1990experimental}. The effect is quantified by subtracting the willingness to pay (the maximum amount one is willing to pay to acquire an object) from the willingness to accept (the minimum amount one is willing to accept to give up the object). \textbf{Gambler's fallacy} delineates the erroneous belief that a departure from what occurs on average will be corrected in the short term, such as believing that a run of one outcome in a chance event will result in an increased probability of the opposite outcome in the next instance. The bias is measured using the questionnaire provided by Leonard and Williams~\cite{leonard2016relationship}. \textbf{Mental accounting} refers to the practice of valuing the same amount of money differently based on subjective criteria, such as the source of the money or its intended use. It is assessed through the task provided by Riger \textit{et al.}~\cite{rieger2022survey}. \textbf{Regression to the mean} describes the tendency to overlook the statistical phenomenon that exceptionally high or low performances or measurements are likely to be followed by more moderate ones. We adopt the task from Toplak \textit{et al.}~\cite{toplak2014assessing} to measure the bias. \textbf{Sunk cost fallacy} is the inclination to continue an endeavor once an investment in money, effort, or time has been made, based on the investment already made rather than current and future costs and benefits. We adopt the task from Bruine de Bruin \textit{et al.}~\cite{bruine2007individual} to measure the fallacy. The degree of rationality refers to the lower susceptibility towards various biases.

An example prompt from the economics domain is shown below.
\begin{tcolorbox}[breakable, enhanced, colback=gray!10, colframe=black, boxrule=0.4pt]
\small
\textbf{Prompt for Loss aversion:}\\
Please act like a participant in this survey and answer the following questions.
Now, imagine you have a choice between the following two options:\\
Option A: A lottery with a 50\% chance of winning \$80 and a 50\% chance of losing \$50. \\
Option B: Zero dollars.\\
Which option would you choose?\\
a. Option A\\
b. Option B
\end{tcolorbox}


\subsection{Game Theory} 

In game theory, rational players are expected to maximize their payoffs based on anticipating others' choices, resulting in a Nash equilibrium, where each player's strategy is the best response to others' strategies. In particular, we choose second-price auction, beauty contest, one-shot prisoner's dilemma, finitely repeated prisoner's dilemma, one-shot public goods game, and finitely repeated public goods game. All these games have a unique subgame perfect Nash equilibrium (SPNE), and we measure rationality by the extent to which agents can play as Nash equilibrium. 

\textbf{Second price auction (SecPriAuc).} This is an auction game with only one item. Two bidders simultaneously bid a price. Then the bidder with a higher price wins the item and only needs to pay the second highest bid. A dominant strategy in this game is to place a bid equaling to the bidder's value of the item. As a result, we calculate the rationality degree as the deviation of bid from one's value: $R=|\text{bid}-\text{value}|/\text{value}$.

\textbf{Beauty contest (BeaCon).} In this game, two players simultaneously choose a number between 0 and 100, and the winner is the player whose number is closest to two-thirds of the average of all chosen numbers. The Nash equilibrium in this game is to choose zero, so the rationality degree is calculated as $R=|100-\text{number chosen}|/100$.

\begin{figure}[t!]
    \centering
    \subfloat[Prisoner's dilemma]{\includegraphics[width=0.32\linewidth]{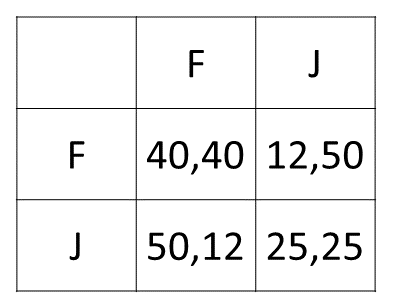}} \ 
    \subfloat[Stag hunt]{\includegraphics[width=0.32\linewidth]{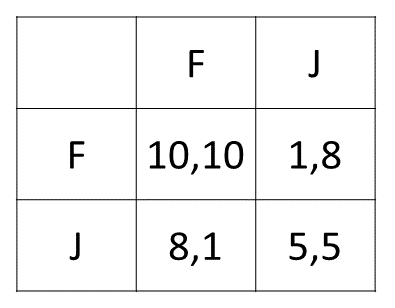}} \ 
     \subfloat[Battle of the sexes]{\includegraphics[width=0.32\linewidth]{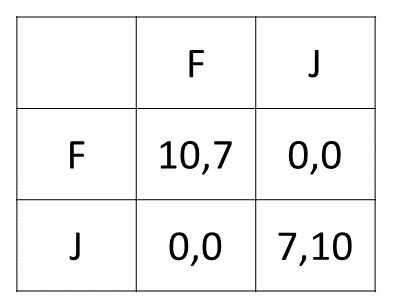}} \ \hspace{0.05cm}
    \caption{Payoff matrix of prisoner's dilemma, stag hunt and battle of the sexes.}
    \label{fig:payoff}
\end{figure}

\textbf{One-shot prisoner's dilemma (OnePD).} This a classic game where two players simultaneously choose to cooperate (choice F) for mutual benefit or defect (choice J) for individual earnings, as shown in Figure~\ref{fig:payoff}(a). The unique Nash equilibrium here is mutual defection, so we calculate the rationality degree as the defection rate.

\textbf{Finitely repeated prisoner's dilemma (FinRepPD).} This is the repeated version of prisoner's dilemma where two players play the prisoner's dilemma for ten rounds. The payoff matrix in each round is the same (Figure~\ref{fig:payoff}(a)). The unique SPNE is that both players defect in every round. Therefore, we measure the rationality degree as the average defection rate of all rounds.

\textbf{One-shot public goods game (OnePG).} In this game, each player has some tokens and simultaneously choose how many of them to contribute to the public account. Then all tokens in the public account are multiplied by a factor and equally shared among all players. Although everyone contributing all tokens to the public can yield the maximum total payoff, the only Nash equilibrium in this game is zero contributions of all players. Therefore, we measure the rationality degree as the percentage of private tokens.

\textbf{Finitely repeated public goods game (FinRepPG).} This is the repeated version of public goods game and the unique SPNE is to keep all tokens as private in all rounds. Similar to one-shot PG, the rationality degree is calculated as the average percentage of private tokens of all rounds.

An example prompt from the game theory domain is shown below.
\begin{tcolorbox}[breakable, enhanced, colback=gray!10, colframe=black, boxrule=0.4pt]
\small
\textbf{Prompt for Beauty Contest:}\\
Please act as a human player. You are playing a game with another human player (called the opponent) for only one round. You and the opponent will simultaneously choose a number between 0 and 100 (including 0 and 100). The winner is the player whose number is closest to two thirds of the average of all chosen numbers. The winner will earn 100 points, and the other player will earn nothing. If there is a tie, both players will earn 50 points.
Your goal is to maximize your own earnings. You should first think carefully about what to do, and then make a decision.
Now let's start the game. Tell me how you think and the number you would like to choose. Please answer in json format with keys 'thought' and 'number'. The 'number' should be between 0 and 100 (including 0 and 100).
\end{tcolorbox}

\subsection{Collective Rationality} 

In the society domain, we measure rationality by cooperation and coordination, and wisdom of crowds.

\textbf{Rationality of cooperation and coordination.} Similar to game theory, this evaluation also focuses on the interaction between agents. However, what we are concerned with here is whether agents can achieve better results together through cooperation and coordination instead of competing with each other. Therefore, we let LLM agents play games that need cooperation (infinitely repeated prisoner's dilemma) and coordination (battle of the sexes, minimum effort, stag hunt). 
In these games, the rationality is measured by the efficiency~\cite{bednar2012behavioral}, which is defined as the actual payoff normalized by the maximum payoff and minimum payoff the player might receive:
\begin{equation}
    \text{Efficiency} = \frac{\text{Payoffs}_{ actual} - \text{Payoffs}_{ min}}{\text{Payoffs}_{ max} - \text{Payoffs}_{ min}}
\end{equation}
This metric lies in $[0,1]$ and a larger efficiency indicates that agents can achieve better payoffs, thus being more rational. Specifically, in the infinitely repeated prisoner's dilemma, cooperation can be supported as a subgame perfect Nash equilibrium with the highest efficiency. In the coordination games, there are multiple Nash equilibria with different efficiency, and rational players are expected to play the Pareto-efficient one.

\textbf{Wisdom of crowds.} This means that groups can make better and more rational decisions than individuals. We let a group of LLM agents collaborate to solve a problem through aggregating the answers~\cite{aher2023using} or debate~\cite{zhang2024exploring}. In the first experiment, we follow~\cite{aher2023using} and ask a group of LLM agents with different profiles to answer 5 general-knowledge questions. In the second experiment, LLM agents are asked to solve multiple-choice~\cite{hendrycks2020measuring} or math problems~\cite{hendrycks2021measuring}, where several LLM agents discuss with each other and revise their answers. The rationality degree is measured by the accuracy of answers.

An example prompt from the collective rationality domain is shown below.
\begin{tcolorbox}[breakable, enhanced, colback=gray!10, colframe=black, boxrule=0.4pt]
\small
\textbf{Prompt for Stag Hunt:}\\
Please act as a human player. You are playing a game with another human player (called the opponent) for 10 rounds.
In each round, you and the opponent will simultaneously choose an action between F and J.
If you choose J and the opponent chooses J, you earn 10 points and the opponent earns 10 points in this round...
Before you choose an action, your actions, the opponent's actions and your earnings in each of the previous rounds will be shown. Your goal is to maximize your own total earnings in all 10 rounds. In each round, you should first think carefully about what to do, and then choose one of the two actions: F or J.
The history of the game is listed as follows delimited by triple backticks.\\
```\\
\{history\}\\
```\\
It is round-\{round\} out of 10 rounds now. Tell me how you think and the action you would like to choose. Please answer in json format with keys 'thought' and 'action'. The 'action' should be F or J. Please return only one json.
\end{tcolorbox}

\section{Supplementary Information for Experiment Setup}

\subsection{Toolkit}\label{sec::toolkit} 
\subsubsection{Input: Configuration and Arguments}
In order to seamlessly test a collection of models on the same prompts without having to change contexts constantly, we develop a toolkit that queries multiple models at a time for a given input prompt and reports the relative and overall performance across all of them. As seen in figure \ref{fig:bench}, the user first passes a prompt that can include both text and image. When submitted, the data will be sent to the API servers of each of the models to retrieve the responses.
After the prompt is sent, the toolkit will create two separate workflows, one requesting the APIs and another requesting the offline models on the local server. 

\subsubsection{Model Response}

\textbf{API server:} The models called via the toolkit include all the models in our work that are accessible by API, namely those that are commercial rather than open-source: gpt-4, gpt-3.5, bard, claude-instant, text-bison, text-davinci-002, and text-davinci-003. The APIs are requested asynchronously so that multiple APIs can be called and their responses received concurrently. 
The models that can accept image inputs are bard and gpt-4, whereas the rest accept exclusively text. 
The API interface that is requested for each model is in some cases is made available by its provider, thereby allowing developers to easily integrate theirs into a versatile array of applications. However, in other cases, third-party APIs are used. Specifically, gpt-4, gpt-3.5, text-davinci-003, and text-davinci-003 are accessed via OpenAI's platform\footnote{https://openai.com}, text-bison is accessed via Google's PaLM API interface\footnote{https://ai.google.dev/models/palm}, bard is queried via a third-party API\footnote{https://github.com/dsdanielpark/Bard-API}, and finally, claude-instant is accessed via Poe\footnote{https://poe.com}.

\textbf{Local server:} Aside from the commercial models, which are accessed by API, there is a bundle of open-sourced models which is widely used. In our benchmark, we provide a general, easy-to-use interface to call the locally deployed models. That is, with the arguments specified for the path of the deployed large language models (or automated downloading from mirror websites when the name of the OSS model is given). The usage of OSS is similar to that of the commercial models based on the uniform interface of both input and output, as described above.
 
\subsubsection{Performance Testing}
Once the responses from both the APIs and the local server are retrieved, the toolkit will automatically collect and process the results according to our benchmarking scheme, which includes each of the forms of rationality tested in this work: psychology, cognitive, decision making, economics, game theory, and collective rationality. Once the rationality performance is gauged for all the models, their results are compared across each of the groupings under each form of rationality tested. For example, under psychology, there are prompts for both theoretical and practical rationality. Under each sub-grouping, there are individual prompts to test different aspects of it, e.g., self-reflection and insight. These results are then passed to the next module for processing and visualization, as described below.

\subsubsection{Output: Results and Comparison}
After all sub-groupings under each rationality type are fully processed, the toolkit will output a heatmap comparing the results across each model. Finally, once all forms of rationality are tested, the toolkit offers the ability to visualize the full model benchmark comparison across all forms of rationality in addition to human-LLM comparison, offering a bird's eye view of the overall state of rationality among the most powerful LLMs. It should be noted that relevant API-keys are required to access certain models, which can be retrieved from the relevant API platforms. Therefore, prior to using the toolkit, one must register for the services on their respective platforms, which in some cases may involve a fee. 

\subsection{Models} 
\label{app:models}

\begin{table*}[h]
\centering
\caption{LLMs utilized in our benchmark.}
\label{table::models}
\small
\renewcommand{\arraystretch}{1.05}
\resizebox{0.7\linewidth}{!}{
\begin{tabular}{|c|c|c|c|c|c|}
\hline
\textbf{Type} & \textbf{Model} & \textbf{Size*} & \textbf{Year} & \textbf{Version} & \textbf{Creator} \\
\hline
\multirow{9}{*}{API} 
& gpt-4o \cite{hurst2024gpt} & $\sim$200B & 2024 & 2024-08-06 & OpenAI \\
& gpt-4 \cite{gpt4openai} & $\sim$1.8T & 2023 & 1106 & OpenAI \\
& gpt-3.5 \cite{gpt35openai} & $\sim$175B & 2022 & 1106 & OpenAI \\
& text-davinci-002 \cite{Ouyang_Davinci_2022} & $\sim$175B & 2022 & - & OpenAI \\
& text-davinci-003 \cite{Ouyang_Davinci_2022} & $\sim$175B & 2022 & - & OpenAI \\
& deepseek-v2.5 \cite{deepseekv2} & 236B & 2024 & v2.5 & DeepSeek \\
& bard \cite{bard_Manyika_Hsiao} & $\sim$137B & 2023 & - & Google \\
& text-bison \cite{Anil_palm2_2023} & $\sim$340B & 2023 & - & Google \\
& claude-instant \cite{Claude} & $\sim$130B & 2023 & v1.2 & Anthropic \\
\hline
\multirow{11}{*}{OSS} 
& Qwen2.5-72B-Instruct \cite{qwen2.5} & 72B & 2024 & v2.5 & Alibaba Cloud \\
& Qwen2.5-32B-Instruct \cite{qwen2.5} & 32B & 2024 & v2.5 & Alibaba Cloud \\
& Qwen2.5-7B-Instruct \cite{qwen2.5} & 7B & 2024 & v2.5 & Alibaba Cloud \\
& Llama-2-13b \cite{touvron2023llama} & 13B & 2023 & chat & Meta \\
& Llama-2-7b \cite{touvron2023llama} & 7B & 2023 & chat & Meta \\
& openchat-13b \cite{Wang_Cheng_Zhan_Li_Song_Liu_2023} & 13B & 2023 & v3.2 & Tsinghua \\
& WizardLM-13B \cite{Xu_Sun_Zheng_Geng_Zhao_Feng_Tao_Jiang_2023} & 13B & 2023 & v1.2 & Microsoft \\
& vicuna-13b \cite{vicuna2023} & 13B & 2023 & v1.5 & LMSYS \\
& vicuna-7b \cite{vicuna2023} & 7B & 2023 & v1.5 & LMSYS \\
& oasst-12b \cite{laion2023} & 12B & 2023 & sft-4 & LAION \\
& chatglm2-6b \cite{glm2024chatglm} & 6B & 2023 & v2 & Tsinghua \\
\hline
\end{tabular}
}
\begin{tabular}{p{10cm}}
\small \footnotesize{*The sizes of some API-based models are actually unknown as they have not been publicized. $\sim$ represents their estimated parameter size. }\\
\end{tabular}
\end{table*}

We have employed a collection of some of the most widely used models, both open-source and commercial. The commercial models are accessible via APIs (as described in section \ref{sec::toolkit}), whereas the open-source models are downloaded and trained individually, making them more versatile and open to modification. However, commercial models typically perform better due to the access to highly valuable data and computing resources that non-profits and academic institutions seldom have. A complete list of model details, including sizes and a brief description, is provided in Table \ref{table::models}.

\textbf{Commercial LLM models:} There are seven models in total: five models released by OpenAI: gpt-4o, gpt-4, gpt-3.5, text-davinci-002, and text-davinci-003; one model released by Deepseek: deepseek v2.5; two models released by Google: bard, text-bison; and one model released by Anthropic: claude-instant. The number of the parameters used to train these models is undisclosed although it can be estimated. 

\textbf{Open-Source LLM models:} There are eight models from a variety of academic and corporate institutions. Notably, llama2-13b and llama2-7b are provided open source by Meta, whereas the others (Qwen-72b/32b/7b, openchat-13b, wizardlm-13b, vicuna-13b, vicuna-7b, oasst-12b, chatglm2-6b) are from academic or other institutions who have released their source code publicly. In contrast to commercial models, they all publicize the number of parameters they use.  

Most of the models follow a similar paradigm of reinforcement with human feedback on top of a pre-trained transformer model. However, the ways in which they differ usually involve fine-tuning with additional or special types of data, as well as their overall number of parameters. Additionally, the way in which humans play a role in the reinforcement learning stage varies across each one.

\section{Supplementary Experiment Results}
\label{sec:supple_results}

\subsection{Correlation Among Different Domains.}
\begin{figure*}[ht]
    \centering
    \subfloat[Correlation among different domains]
    {\includegraphics[width=0.45\linewidth]{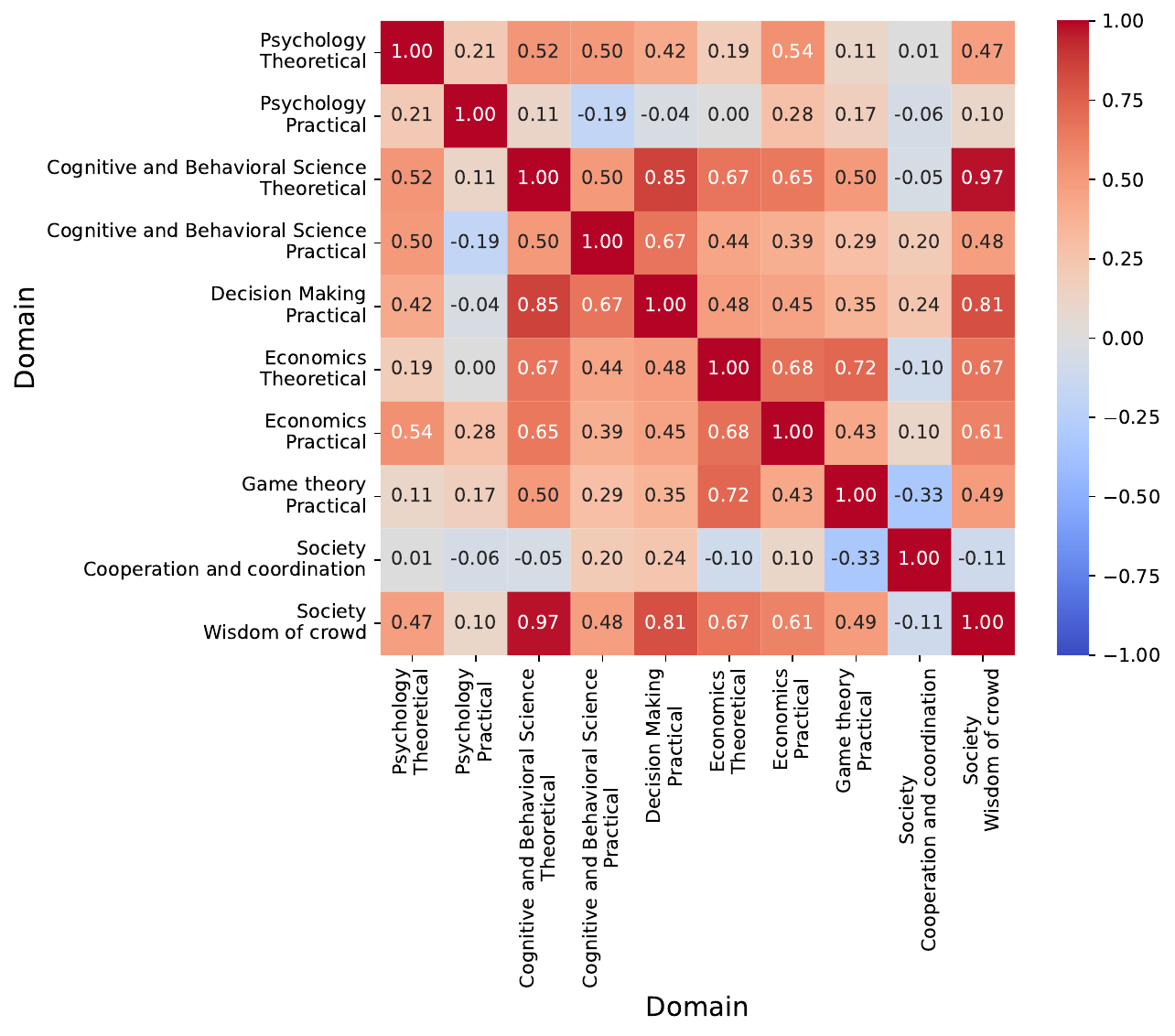}}  \
    \subfloat[Correlation among different domains (practical and theoretical rationality is merged).] 
    {\includegraphics[width=0.45\linewidth]{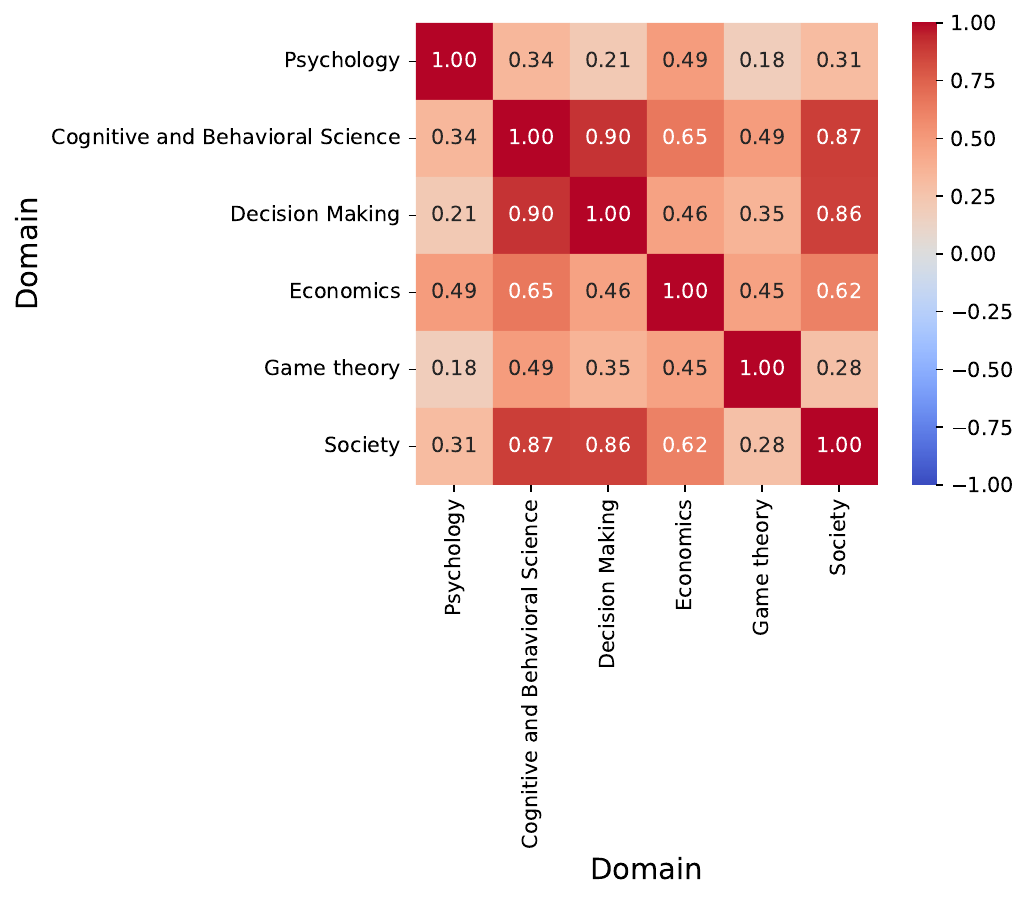}} \
        \caption{The correlation of the average rationality in different domains.}
    \label{fig:corr}
\end{figure*}

\begin{figure*}[t]
    \centering
     \subfloat[Impact of parameter size]{
    \includegraphics[width=0.48\linewidth]{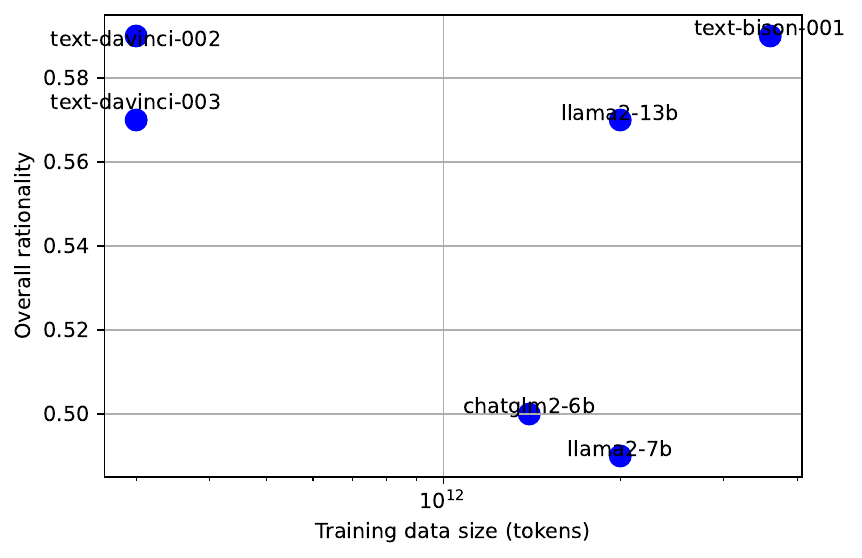}}\
           \subfloat[Impact of dataset size]{\includegraphics[width=0.48\linewidth]{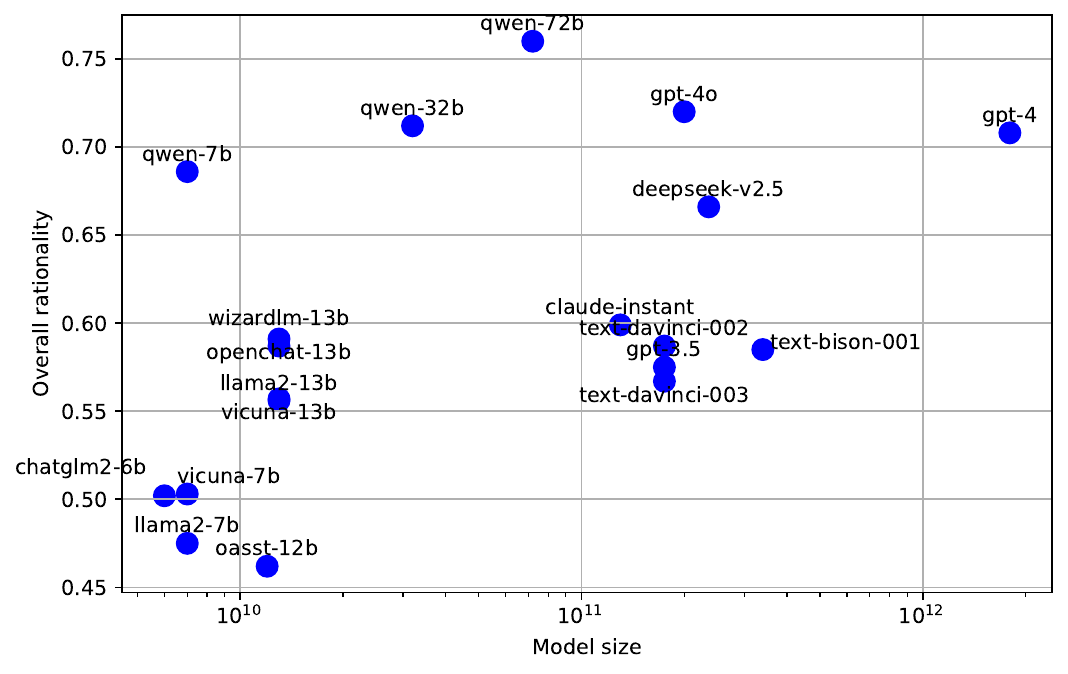}}\
            \caption{Impact of parameter size and data size on rationality.}
    \label{fig:size}
\end{figure*}

The evaluation of rationality is conducted on different domains, leaving us an interesting question of how the ability in different domains is correlated.
That is, some domains' rationality may share a similar definition, or the ability level depends on some common abilities.
Therefore, to present the correlation among different domains, we present the results in Appendix Figure~\ref{fig:corr}. Specifically, in Appendix Figure~\ref{fig:corr} (b), we further merge the theoretical and practical evaluations in each domain by computing the average. From the results, we have the following observations.

From the overall view, rationality in different domains is highly related. In these two heatmap figures, most of the correlation values are positive, revealing strong relations in different domains. These can be explained in two ways. First, many different scales in various domains are similar, with different emphases, but they share the same ability. Second, the abilities affecting the rationality level may depend on some shared basic abilities. For example, in theoretical rationality, the reasoning ability may affect a lot of aspects of rationality in different domains.

In addition, some domains have significantly lower correlations than others. Specifically, for ``psychology-practical'', we can observe that its correlation is much smaller, which can be explained by the fact that EQR tests in psychology-practical focus more on the quite different aspects of emotional regulation compared with scales in other domains or the theoretical rationality of psychology domain.

\subsection{Impact of Dataset Size and Model Size}
LLMs exhibit remarkable abilities that correlate with their parameter sizes~\cite{zhao2023survey}. Thus, studying how model size affects rationality is crucial. Additionally, the training dataset size influences their reasoning and knowledge, prompting an examination of its impact on rationality evaluations. Appendix Figure~\ref{fig:size} shows a strong positive correlation between model size and rationality, with larger models generally displaying higher rationality. However, no strong relationship is observed between rationality and training dataset size, which may be influenced by factors such as dataset quality, varying training methodologies, and the sufficiency of data used in training.




\subsection{Impact of Reinforcement Learning with Human Feedback}

\begin{figure*}[h!]
    \centering
    \includegraphics[width=0.50\linewidth]{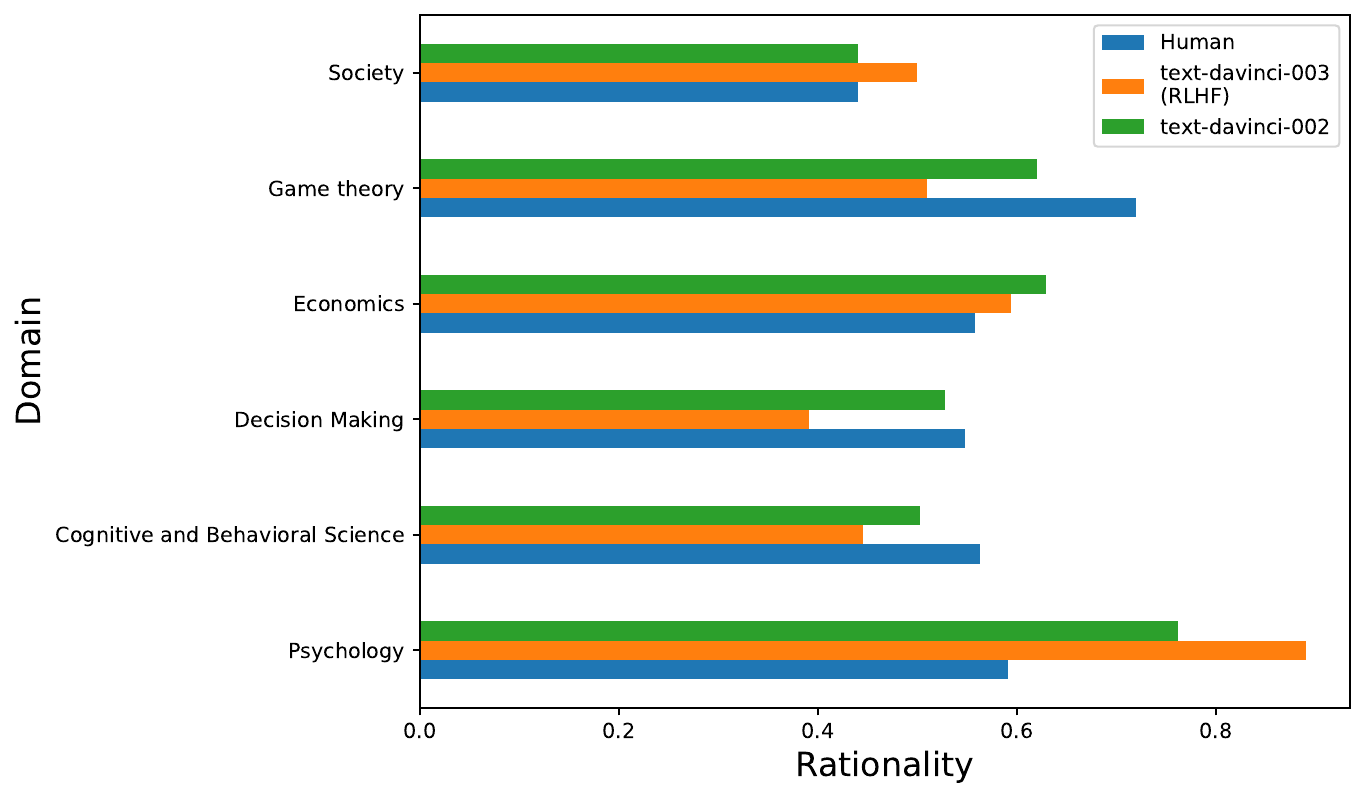}
    \includegraphics[width=0.40\linewidth]{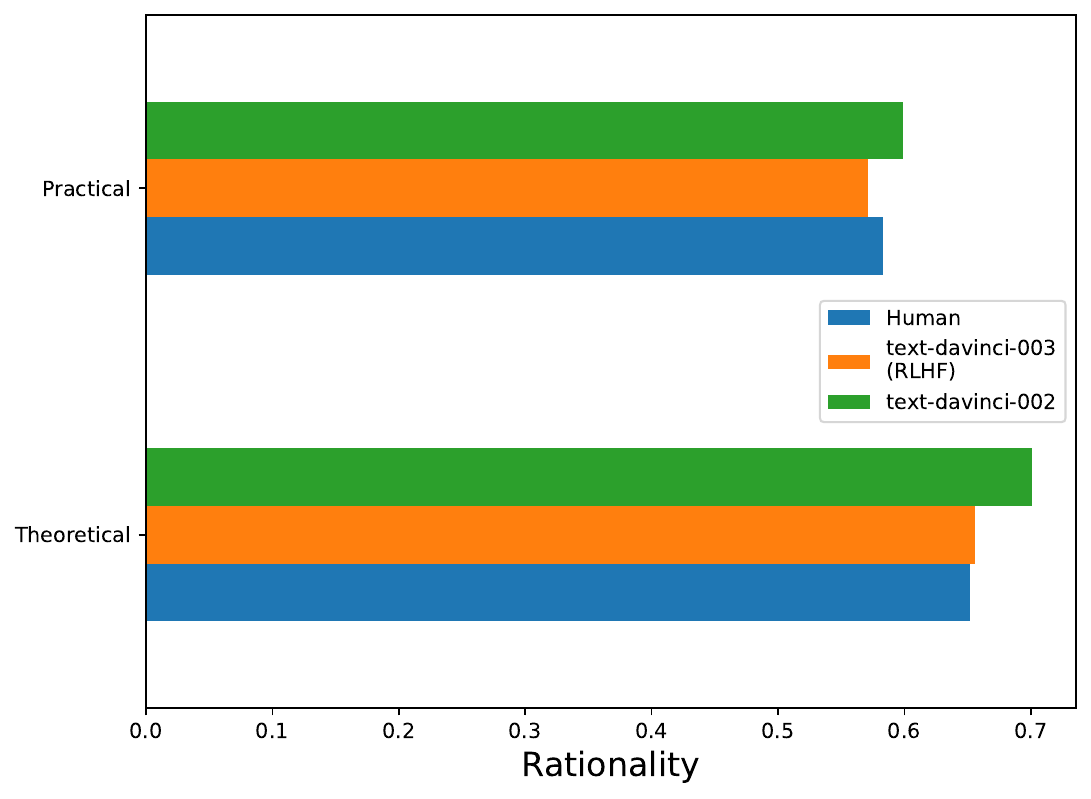}
          \caption{Impact of reinforcement learning with human feedback.}
    \label{fig:rlhf}
\end{figure*}

\begin{figure*}[h]
    \centering
    \includegraphics[width=0.65\linewidth]{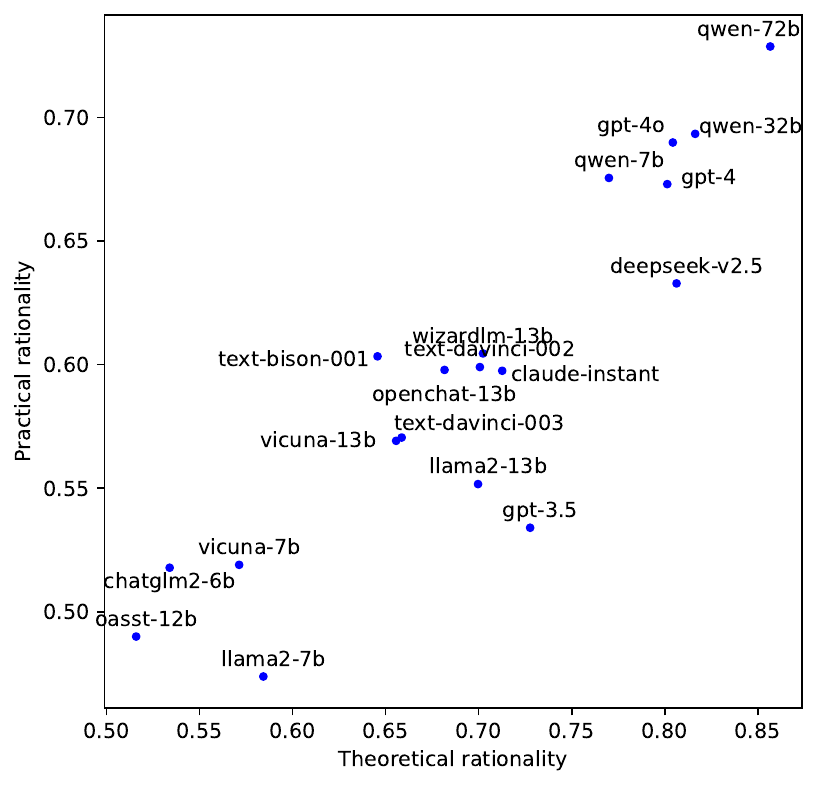}
    \caption{Relation between theoretical rationality and practical rationality.}
    \label{fig:theo-prac}
\end{figure*}

Reinforcement learning with human feedback (RLHF)~\cite{ouyang2022training} is the recent advance of reinforcement learning, which has shown good performance in training language models. Specifically, it is different from the traditional reinforcement learning methods, which receive the reward from the environment, such as the score in the game environment; instead, in RLHF, the model can take feedback from humans. Therefore, in training LLMs, the human feedback integrated by RLHF also takes human characteristics into the model. One representative example is the value alignment in politics, wherein ChatGPT has similar preferences with those about forty hired men/women in labeling the training samples in RLHF.
As a result, when assessing whether the LLMs behave rationally, RLHF is worth studying as it may cause the LLM to exhibit a similar amount of high or low rationality as real human beings.
Since text-davinci-003 is the improved version of text-davinci-002, having introduced the technique of RLHF, comparing rationality between the two can help us understand the impact of RLHF, for which the results are shown in Appendix Figure~\ref{fig:rlhf}.


\subsection{Correlation Between Theoretical and Practical Domain.}
As mentioned above, rationality can be overall divided into theoretical rationality and practical rationality, of which the previous one focuses more on how to reason, and the latter about whether the action is rational. As illustrated in Appendix Figure~\ref{fig:theo-prac}, there is a strong correlation between theoretical rationality and practical rationality. GPT-4 has both the best practical rationality and practical rationality overall. Those open-source models with the fewest model parameters have the lowest theoretical and practical rationality. GPT-3.5 has high theoretical rationality but low practical rationality, which is very interesting. As we know, theoretical rationality is more about basic reasoning, belief, and thinking, while practical rationality ensures the selected action is rational given the belief is correct.




\subsection{Data Contamination Analysis}

One potential limitation of our work is the data contamination of LLMs. The contamination problem arises when LLMs have previously encountered similar rationality questionnaires during their training, which could lead to an overestimation of their rationality. Essentially, the LLMs might be recalling answers rather than genuinely reasoning through the questions, thus skewing the results of our benchmarking efforts. Such data contamination problem is widely recognized as a universal issue in truthful LLM evaluation~\cite{deng2023investigating,golchin2023time}, especially as LLM are gaining increasing access to enormous training datasets.

In this section, we conduct experiments to assess the potential impact of data contamination issue.
Specifically, for experiments at the interpersonal and societal level, the payoff matrices of games and prompts are manually constructed, and thus there is no data contamination issue. However, the questions in the individual level are from existing scales and tests. To verify whether data contamination affects the results, we slightly modify the questions of several tests to ensure that the questions do not appear in the training data of LLMs. Then, we test different LLMs on the modified questions, and compare the results with original ones.

Specifically, we choose three typical tests, including the Cognitive Reflection Test, Base-Rate Neglect (Statistical), and Conjunction Fallacy. These tests are popularly used and thus are likely to have data contamination issues. We modify these tests from the following aspects.

\textbf{Changing character names.} For example, for the following question in Base-Rate Neglect (Statistical) test, we replace the character name ''John'' with a randomly sampled name~\cite{aher2023using} without changing the answers as follows:

\begin{tcolorbox}[breakable, enhanced, colback=gray!10, colframe=black, boxrule=0.4pt]
\small
\textbf{Original:}\\
\textit{Among the 1000 people that participated in the study, there were 995 nurses and 5 doctors. \textbf{John} is randomly chosen participant in this research. He is 34 years old. He lives in a nice house in a fancy neighborhood. He expresses himself nicely and is very interested in politics. He invests a lot of time in his career. Which is more likely?\\
  a. \textbf{John} is a nurse.\\
  b. \textbf{John} is a doctor.}

\noindent\textbf{Modified:}\\
\textit{Among the 1000 people that participated in the study, there were 995 nurses and 5 doctors. \textbf{Mr. WERITO} is randomly chosen participant in this research. He is 34 years old. He lives in a nice house in a fancy neighborhood. He expresses himself nicely and is very interested in politics. He invests a lot of time in his career. Which is more likely?\\
  a. \textbf{Mr. WERITO} is a nurse.\\
  b. \textbf{Mr. WERITO} is a doctor.}
\end{tcolorbox}

\textbf{Changing the numerical value in the questions.} For example, for the following question in Cognitive Reflection Test, we change the number “48” to “24”, and the correct answer also changes from 47 to 23.

\begin{tcolorbox}[breakable, enhanced, colback=gray!10, colframe=black, boxrule=0.4pt]
\small
\textbf{Original:}\\
\textit{In a lake, there is a patch of lily pads. Every day, the patch doubles in size. If it takes \textbf{48} days for the patch to cover the entire lake, how long would it take for the patch to cover half of the lake? \_ days}

\noindent\textbf{Modified:}\\
\textit{In a lake, there is a patch of lily pads. Every day, the patch doubles in size. If it takes \textbf{24} days for the patch to cover the entire lake, how long would it take for the patch to cover half of the lake? \_ days}
\end{tcolorbox}

Considering that LLMs from different creators may use different training datasets, we conduct experiments on LLMs from a wide range of creators, including OpenAI, Meta, Deepseek, Alibaba, Tsinghua, Microsoft, LMSYS, and LAION. Their rationality scores are presented in Table~\ref{table:data_contamination}, calculated as the accuracy on these questions. It can be observed that most LLMs obtain similar rationality scores after modifying the questions, indicating that the data contamination issue does not affects the results significantly.

\begin{table}[h]
\centering
\caption{Comparison of rationality scores on modified questions with original questions.}
\label{table:data_contamination}
\resizebox{0.7\linewidth}{!}{
\begin{tabular}{c|cc}
\hline
                       & \textbf{Original} & \textbf{Modified} \\ \hline
\textbf{gpt-4o}        & 0.882             & 0.882             \\
\textbf{deepseek-v2.5} & 0.647             & 0.588             \\
\textbf{openchat-13b}  & 0.294             & 0.294             \\
\textbf{wizardlm-13b}  & 0.176             & 0.118             \\
\textbf{vicuna-13b}    & 0.176             & 0.118             \\
\textbf{oasst-12b}     & 0.235             & 0.353             \\
\textbf{qwen-7b}       & 0.706             & 0.706             \\
\textbf{llama-7b}      & 0.118             & 0.059             \\ \hline
\end{tabular}
}
\end{table}

\subsection{Comparison with Reasoning Model}
In this section, we test several recent reasoning models on our benchmark, including OpenAI o1, Deepseek R1, and DeepSeek-R1-Distill-Qwen-32B (a distilled model based on Qwen2.5-32B using outputs from DeepSeek R1).
As shown in Table~\ref{table:reasoning_model}, reasoning models such as o1 show a higher overall rationality score compared to GPT-4o. However, Deepseek R1 has a slightly lower score than Deepseek V3, while DeepSeek-R1-Distill-Qwen-32B outperforms Qwen-32B. Deepseek V3 holds the highest cognitive rationality score among the LLMs. This suggests that while reasoning models can demonstrate high rationality, they do not always lead to more cognitively rational behavior. Based on our observations, these models tend to spend significantly more time in their reasoning processes, sometimes leading to overthinking during their responses. This pattern of reasoning models not consistently outperforming other models is evident across the various domains analyzed. 

\begin{table*}[h]
\centering
\caption{Comparison of rationality score with reasoning models on cognitive science domain.}
\label{table:reasoning_model}
\resizebox{0.99\linewidth}{!}{
\begin{tabular}{|l|l|c|c|c|c|c|c|c|}
\hline
\textbf{Aspect} & \textbf{Scale} & \textbf{Human} & \textbf{o1} & \textbf{GPT-4o} & \textbf{Deepseek-R1} & \textbf{Deepseek-V3} & \textbf{Deepseek-R1 Distill} & \textbf{Qwen-32b} \\
\hline
\multirow{3}{*}{\textbf{Dual System Thinking}} 
& REI Rationality & 0.60 & 0.84 & 0.86 & 0.74 & 0.96 & 0.69 & 1.00 \\
& REI Experimentality & 0.37 & 0.28 & 0.28 & 0.20 & 0.21 & 0.23 & 0.26 \\
& Cognitive Reflection Test & 0.21 & 1.00 & 1.00 & 1.00 & 1.00 & 1.00 & 1.00 \\
\hline
\textbf{Inductive Reasoning} & Letter Sets Test & 0.64 & 0.93 & 0.50 & 0.87 & 0.93 & 0.70 & 0.57 \\
\hline
\textbf{Deductive Reasoning} & Logiqa 2.0 & 0.84 & 0.79 & 0.69 & 0.79 & 0.79 & 0.77 & 0.72 \\
\hline
\textbf{Causal Reasoning} & Causal Reasoning & 0.82 & 0.85 & 0.52 & 0.98 & 0.92 & 0.69 & 0.44 \\
\hline
\textbf{Scientific Reasoning} & Scientific Reasoning Scale & 0.62 & 1.00 & 1.00 & 1.00 & 0.91 & 1.00 & 0.91 \\
\hline
\textbf{Defeasible Reasoning} & Defeasible Reasoning & 0.56 & 0.67 & 0.50 & 0.50 & 0.67 & 0.17 & 0.167 \\
\hline
\multirow{2}{*}{\textbf{Deontic Reasoning}} 
& Wason Selection Task - Abstract & 0.19 & 1.00 & 0.33 & 1.00 & 1.00 & 1.00 & 0.33 \\
& Wason Selection Task - Deontic & 0.44 & 1.00 & 1.00 & 1.00 & 1.00 & 1.00 & 1.00 \\
\hline
\multirow{2}{*}{\textbf{Thinking Disposition}} 
& Critical Thinking Disposition Scale & 0.72 & 0.86 & 0.80 & 0.80 & 0.84 & 0.93 & 0.8625 \\
& Actively Open-Minded Thinking Scale & 0.72 & 0.73 & 0.71 & 0.72 & 0.71 & 0.79 & 0.71 \\
\hline
\multirow{5}{*}{\textbf{Heuristics and Biases}} 
& Belief Bias in Syllogistic Reasoning & 0.63 & 0.75 & 0.50 & 1.00 & 1.00 & 1.00 & 0.50 \\
& Bias Blind Spot & 0.75 & 0.72 & 0.63 & 0.74 & 0.76 & 0.80 & 0.41 \\
& Hindsight Bias & 0.80 & 1.00 & 1.00 & 1.00 & 1.00 & 1.00 & 1.00 \\
& Illusion of Control & 0.59 & 0.00 & 0.50 & 0.00 & 0.00 & 0.50 & 1.00 \\
& Regret Aversion & 0.33 & 0.67 & 0.67 & 0.33 & 0.67 & 0.33 & 0.67 \\
\hline
\textbf{Overall} & Overall & 0.58 & 0.77 & 0.68 & 0.75 & 0.79 & 0.74 & 0.68 \\
\hline
\end{tabular}
}
\end{table*}

\subsection{Original Rationality Scores}
\label{app:original_rationality_scores}
\begin{figure}[h]
    \centering
    \subfloat[Theoretical rationality]{\includegraphics[width=\linewidth]{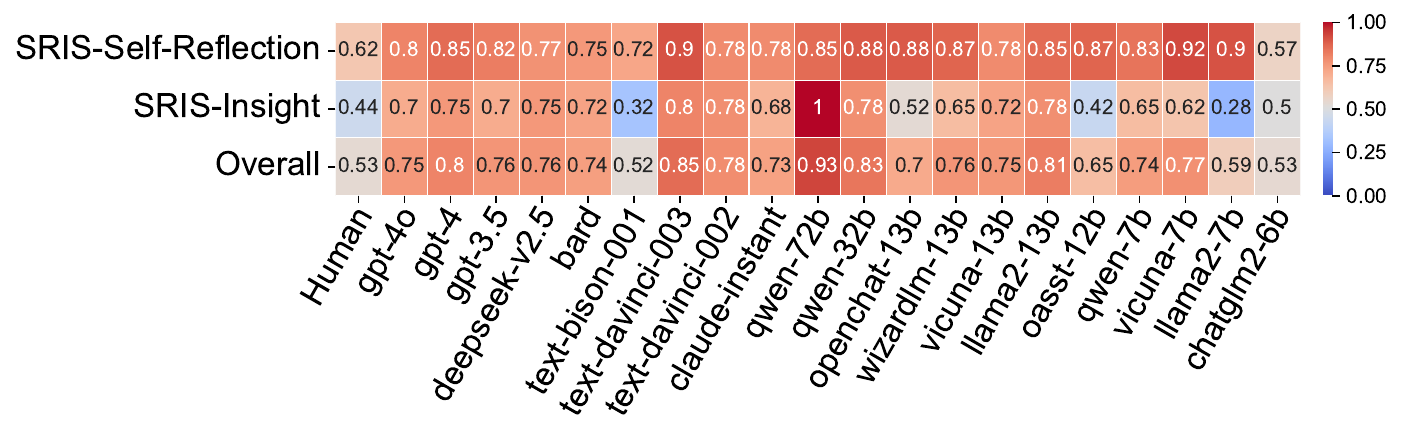}} \ \vspace{0.1cm}
      \subfloat[Practical rationality]{\includegraphics[width=\linewidth]{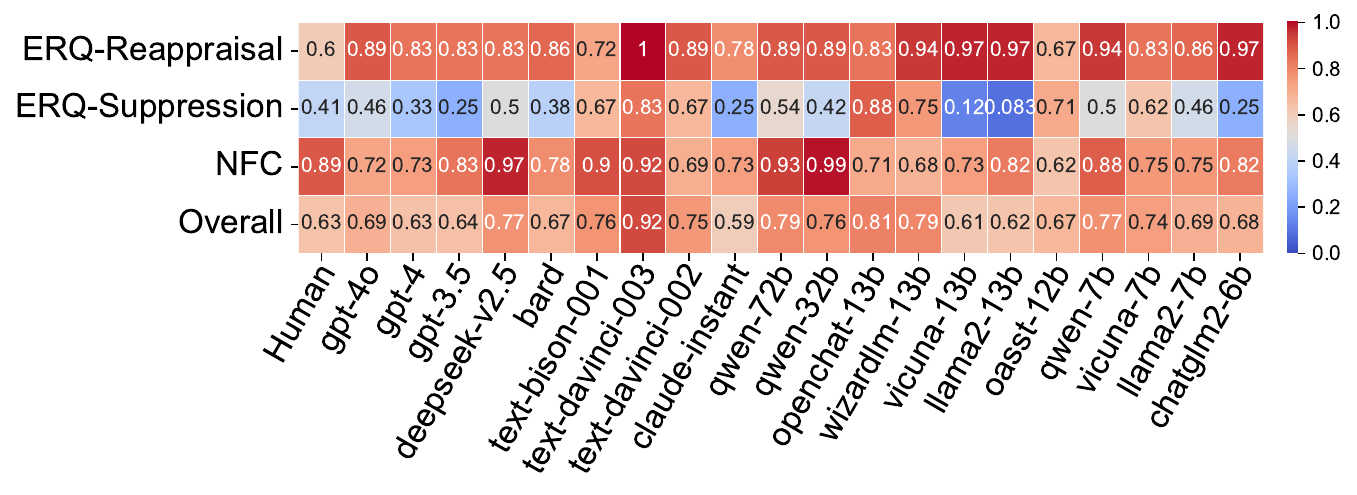}} \ \vspace{0.1cm}
    \caption{Rationality scores of LLMs (x) on different aspects (y) in psychology domain.}
    \label{fig:psy_orig}
\end{figure}

\begin{figure}[h]
    \centering
    \subfloat[Theoretical rationality]{\includegraphics[width=\linewidth]{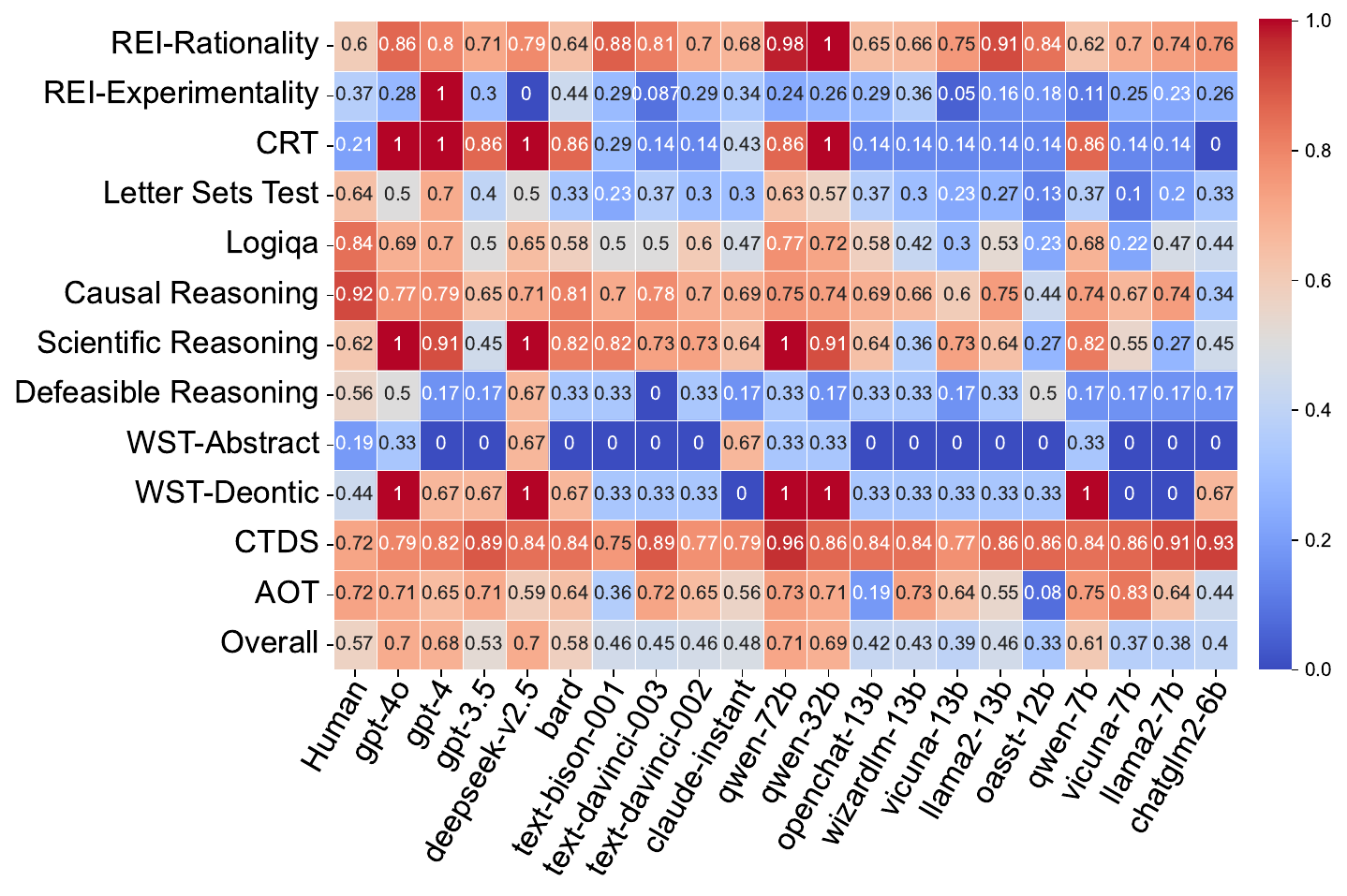}} \ \vspace{0.1cm}
      \subfloat[Practical rationality]{\includegraphics[width=\linewidth]{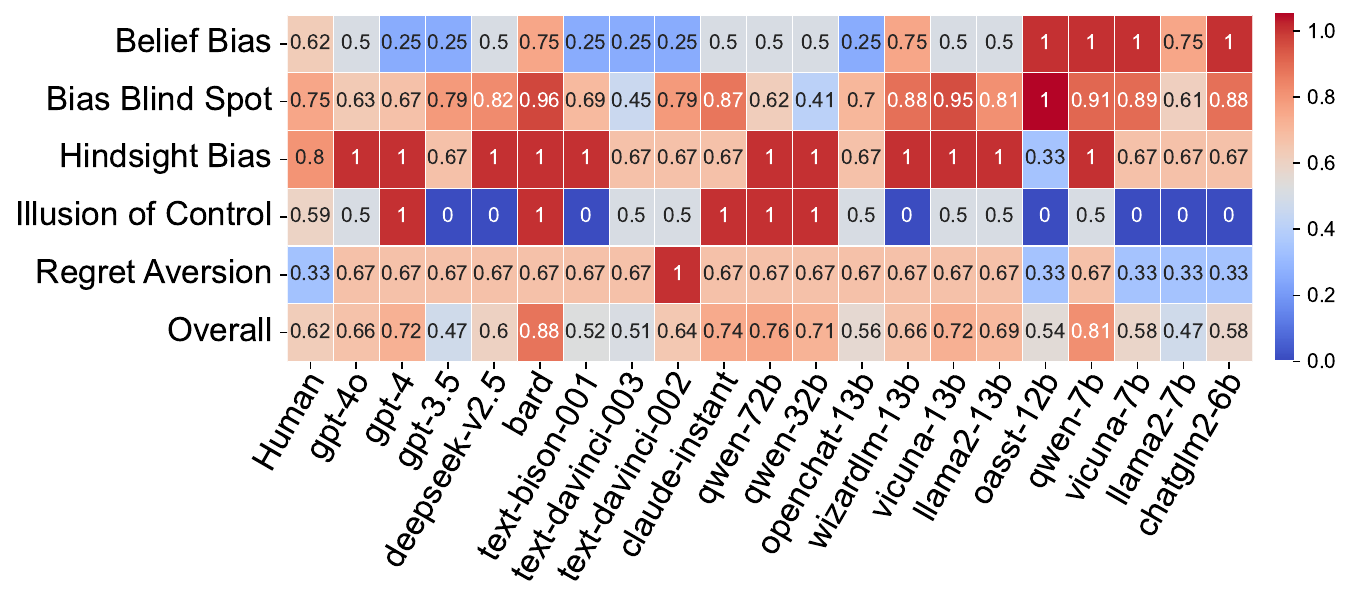}} \ \vspace{0.1cm}
    \caption{Rationality scores of LLMs (x) on different aspects (y) in the cognitive science domain.}
    \label{fig:cog_orig}
\end{figure}

\begin{figure}[h]
    \centering\includegraphics[width=0.9\linewidth]{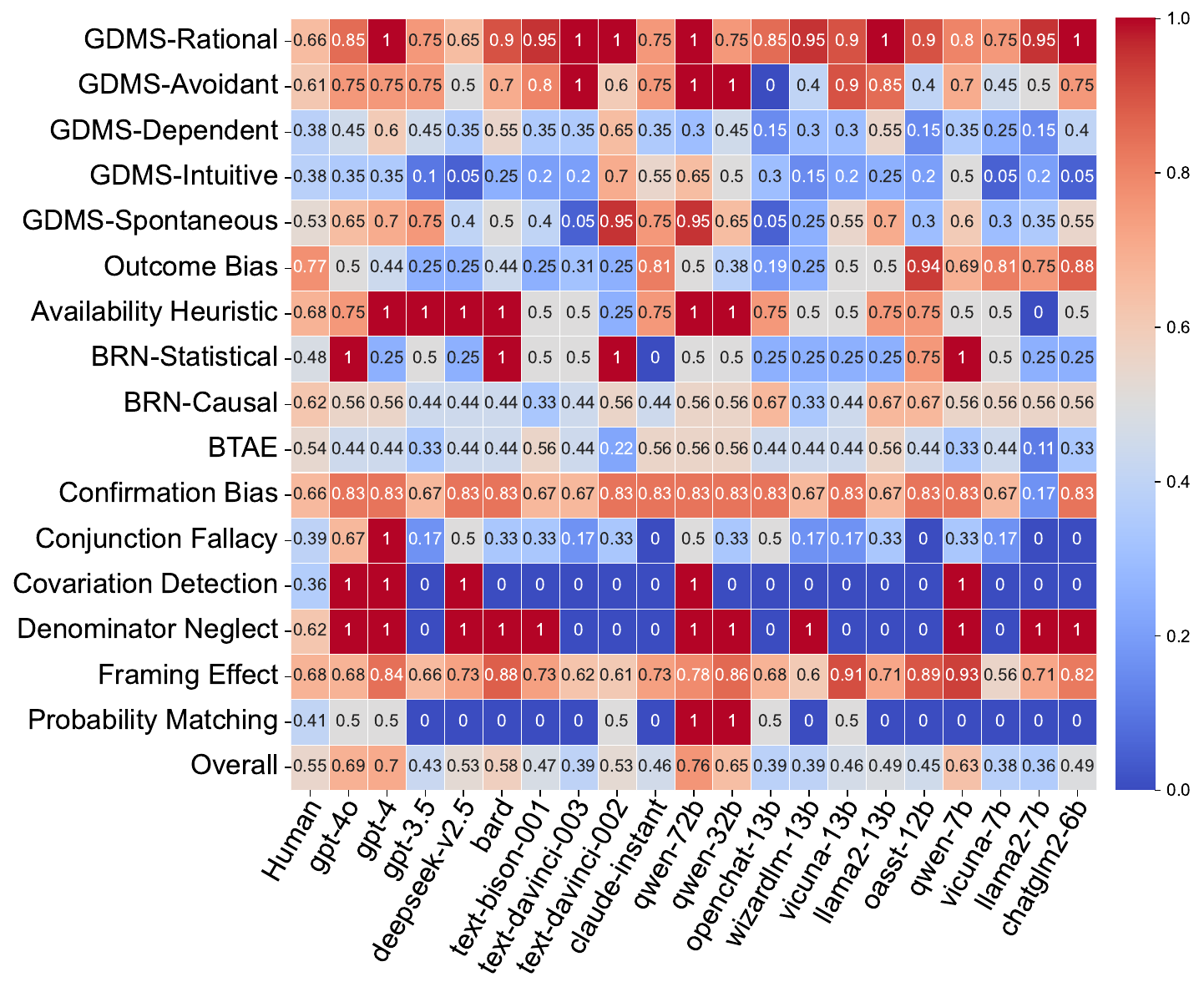}
    \caption{Rationality scores of LLMs (x) on different aspects (y) in decision-making domain.}
    \label{fig:dec_orig}
\end{figure}

\begin{figure}[h]
    \centering
    \includegraphics[width=\linewidth]{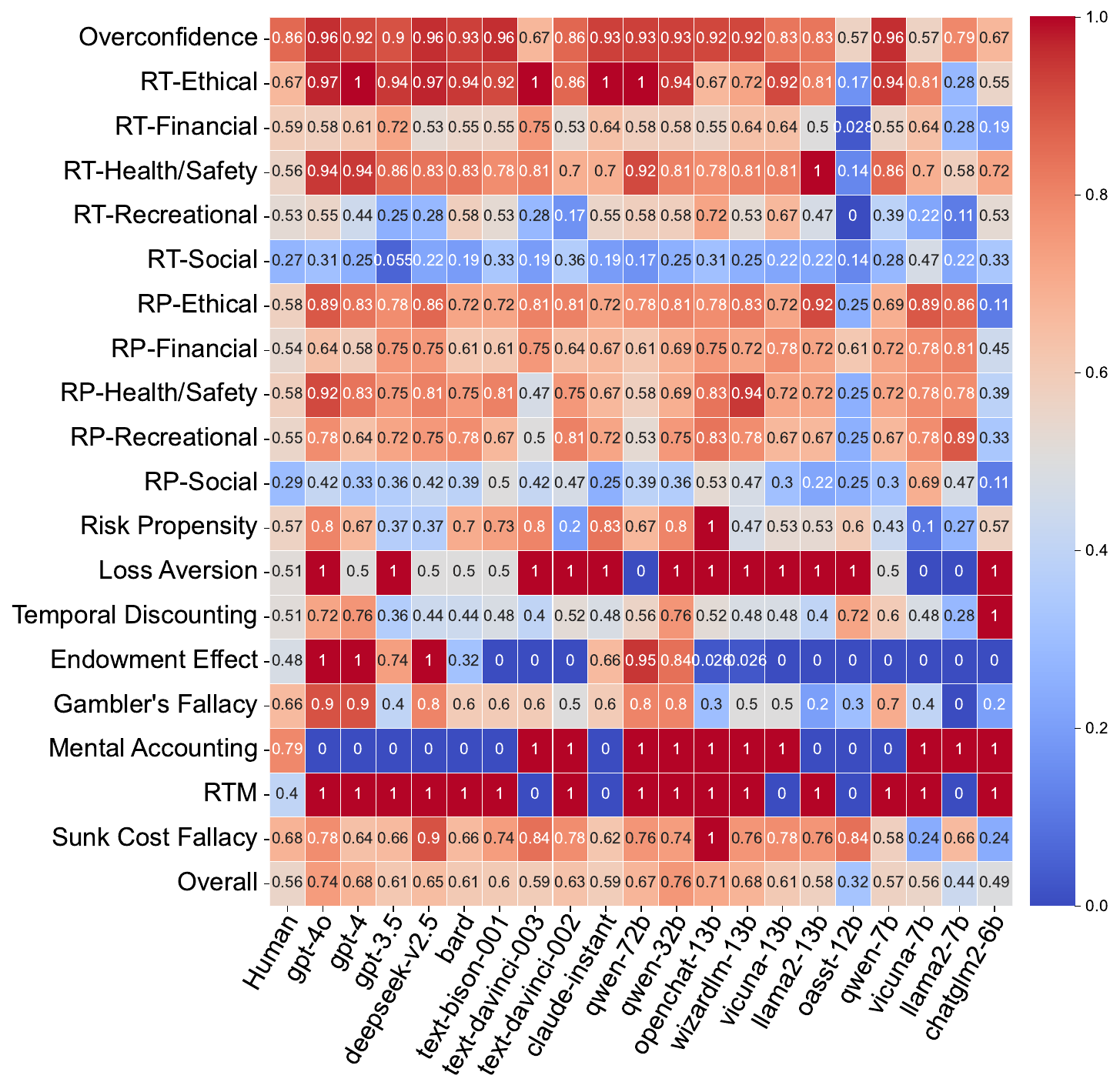}
    \caption{Rationality scores of LLMs (x) on different aspects (y) in economics domain.}
    \label{fig:econ_orig}
\end{figure}

\begin{figure}[h]
    \centering
    \includegraphics[width=\linewidth]{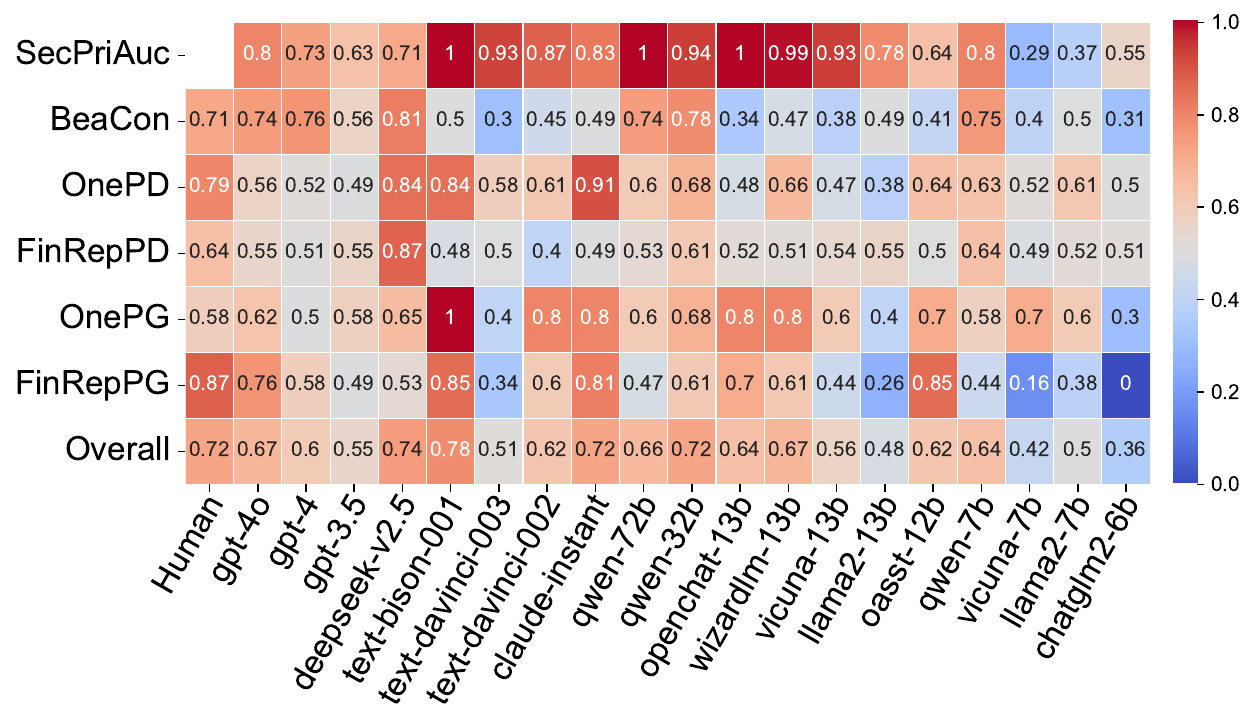}
    \caption{Rationality scores of LLMs (x) on different aspects (y) in game theory domain. (The missing values in the first row is due to the lack of human experiment data of the Second Price Auction game.)}
    \label{fig:game_orig}
\end{figure}

\begin{figure}[h]
    \centering
    \subfloat[Cooperation and coordination]{\includegraphics[width=\linewidth]{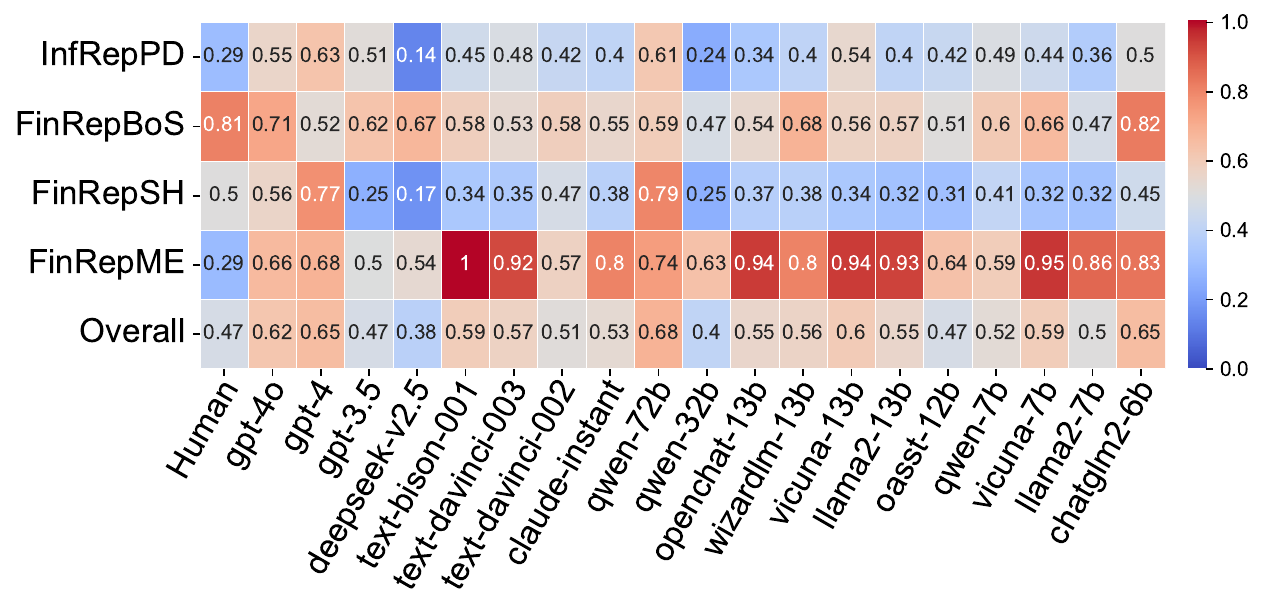}} \ \vspace{0.02cm}
      \subfloat[Wisdom of crowds - MMLU and MATH]{\includegraphics[width=\linewidth]{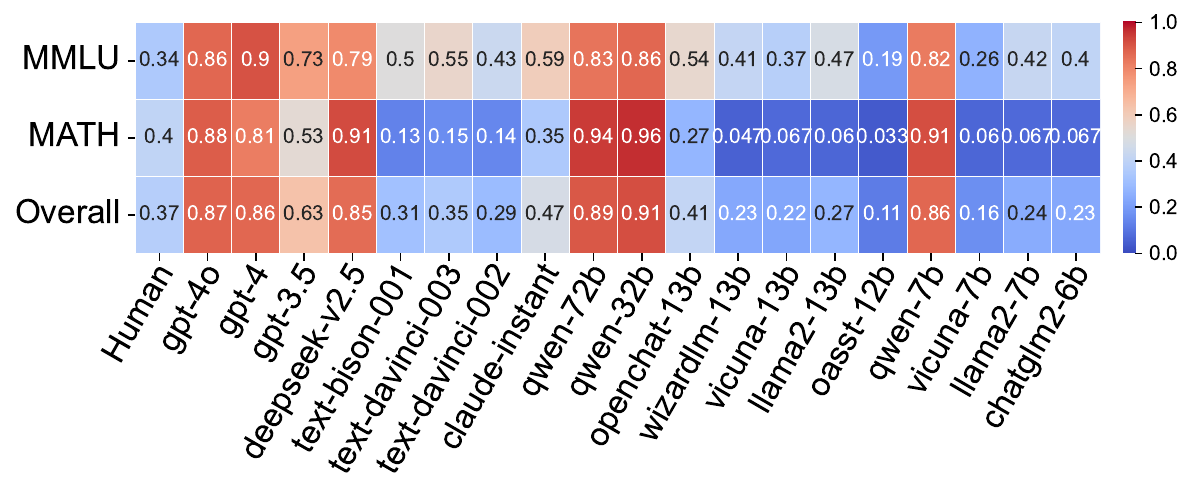}} \ \vspace{0.02cm}
    \caption{Rationality scores of LLMs (x) on different aspects (y) in collective rationality domain.}
    \label{fig:social_orig}
\end{figure}

We present the original rationality scores of LLMs in Figure~\ref{fig:psy_orig},~\ref{fig:cog_orig},~\ref{fig:dec_orig},~\ref{fig:econ_orig},~\ref{fig:game_orig}, and~\ref{fig:social_orig}.

\section{Open Discussions}
\textbf{LLMs as Simulators:} Evaluating LLMs reveals their capacity to emulate human behavior, offering insights that refine their contextual understanding, coherence, and responsiveness. These evaluations allow developers to tailor LLMs for diverse communication scenarios, enhancing their roles in natural language processing, conversational AI, and decision-making. This process ultimately advances LLMs as sophisticated tools that mirror and augment human-like rationality.

\textbf{LLMs as AI Assistants:} Evaluating LLMs enhances their utility in decision-making by leveraging their rationality. These comprehensive evaluations guide the integration of LLMs into decision-making frameworks, aiding in complex scenarios. Insights from these evaluations ensure the reliability and consistency of LLM outputs, making them valuable support systems. This understanding enables decision-makers to improve their judgment, particularly in areas requiring rapid data analysis and nuanced comprehension, fostering a synergistic relationship between human cognition and machine intelligence for more effective decision-making.

\textbf{Rational Models and Predictability:} Measuring LLM rationality is crucial for predicting their behavior in human interactions. Theory of Mind (ToM)~\cite{leslie2004core}, the ability to attribute mental states to oneself and others, is key to this. While LLMs can emulate ToM, it's uncertain if they truly grasp mental states~\cite{ullman2023large,kosinski2023theory}. ToM enables LLMs to understand nuances in human interaction, including sarcasm, humor, and indirect speech~\cite{wang2021towards}. Genuine ToM capability enhances LLMs' rational and contextually appropriate responses. Incorporating ToM into LLMs can develop empathy and moral reasoning, crucial in emotional support or ethical dilemmas. This represents advanced rationality, a key aspect of human intelligence~\cite{cuzzolin2020knowing}. However, integrating ToM into LLMs poses challenges, requiring advanced algorithms and an understanding of human cognitive and social processes. Measuring this capability involves exploring how humans develop ToM~\cite{astington1995theory} and simulating these processes in LLMs. These efforts can lead to AI systems capable of genuinely human-like understanding and interactions.






\end{document}